\documentclass[conference]{IEEEtran}
\IEEEoverridecommandlockouts
\usepackage{cite}
\usepackage{amsmath,amssymb,amsfonts}
\usepackage{algorithmic}
\usepackage{graphicx}
\usepackage{textcomp}
\usepackage{xcolor}
\def\BibTeX{{\rm B\kern-.05em{\sc i\kern-.025em b}\kern-.08em
    T\kern-.1667em\lower.7ex\hbox{E}\kern-.125emX}}

\usepackage{listings}
\usepackage{tabularx}
\usepackage{hyperref}

\usepackage{algorithm}
\usepackage{xcolor}
\usepackage{amsmath}

\newcommand*{\mybox}[2]{\colorbox{#1!30}{\parbox{0.97\linewidth}{#2}}}
\newtheorem{observation}{Highlight}

\usepackage{makecell}
\usepackage{url}
\usepackage{subcaption}
\usepackage{multirow}

\usepackage{booktabs}
\usepackage{tagging}
\usepackage{ipdps26repro}

\usetag{explanation}
\usetag{example}

\begin{document}

\title{Clust-PSI-PFL: A Population Stability Index Approach for Clustered Non-IID Personalized Federated Learning\\

\thanks{Daniel M.\ Jimenez-Gutierrez was partially supported by PNRR351
TECHNOPOLE -- NEXT GEN EU Roma Technopole -- Digital Transition,
FP2 -- Energy transition and digital transition in urban regeneration and
construction. 
Aris Anagnostopoulos was supported by the PNRR
MUR project PE0000013-FAIR, the PNRR MUR project IR0000013-
SoBigData.it, and the MUR PRIN project 2022EKNE5K ``Learning in
Markets and Society.'' 
Ioannis Chatzigiannakis was supported by PE07-
SERICS (Security and Rights in the Cyberspace) -- European Union
Next-Generation-EU-PE00000014 (Piano Nazionale di Ripresa e Re-
silienza -- PNRR). 
Andrea Vitaletti was supported by the project SERICS
(PE00000014) under the MUR National Recovery and Resilience Plan
funded by the European Union - NextGenerationEU.}
}

\author{\IEEEauthorblockN{Daniel M. Jimenez-Gutierrez}
\IEEEauthorblockA{\textit{Department of Computer,} \\
\textit{Control and Management Engineering} \\
\textit{Sapienza University of Rome}\\
Rome, Italy \\  
jimenezgutierrez@diag.uniroma1.it}
\and
\IEEEauthorblockN{Mehrdad Hassanzadeh}
\IEEEauthorblockA{\textit{Department of Computer,} \\
\textit{Control and Management Engineering} \\
\textit{Sapienza University of Rome}\\
Rome, Italy \\
hassanzadeh.1961575@studenti.uniroma1.it}
\and
\IEEEauthorblockN{David Solans}
\IEEEauthorblockA{\textit{Telef\'onica Research}\\
Barcelona, Spain \\
david.solansnoguero@telefonica.com}
\and
\IEEEauthorblockN{Mohammed Elbamby}
\IEEEauthorblockA{\textit{Telef\'onica Research}\\
Barcelona, Spain \\
mohammed.elbamby@telefonica.com}
\and
\IEEEauthorblockN{Nicolas Kourtellis}
\IEEEauthorblockA{\textit{Keysight AI Labs}\\
Barcelona, Spain \\
nkourtellis@gmail.com}
\and
\IEEEauthorblockN{Aris Anagnostopoulos}
\IEEEauthorblockA{\textit{Department of Computer,} \\
\textit{Control and Management Engineering} \\
\textit{Sapienza University of Rome}\\
Rome, Italy \\
aris@diag.uniroma1.it}
\and
\IEEEauthorblockN{Ioannis Chatzigiannakis}
\IEEEauthorblockA{\textit{Department of Computer,} \\
\textit{Control and Management Engineering} \\
\textit{Sapienza University of Rome}\\
Rome, Italy \\
ichatz@diag.uniroma1.it}
\and
\IEEEauthorblockN{Andrea Vitaletti}
\IEEEauthorblockA{\textit{Department of Computer,} \\
\textit{Control and Management Engineering} \\
\textit{Sapienza University of Rome}\\
Rome, Italy \\
vitaletti@diag.uniroma1.it}
}

\maketitle

\begin{abstract}
Federated learning (FL) supports privacy-preserving, decentralized machine learning (ML) model training by keeping data on client devices. However, non-independent and identically distributed (non-IID) data across clients biases updates and degrades performance. To alleviate these issues, we propose Clust-PSI-PFL, a clustering-based personalized FL framework that uses the Population Stability Index (PSI) to quantify the level of non-IID data. We compute a weighted PSI metric, $WPSI^L$, which we show to be more informative than common non-IID metrics (Hellinger, Jensen–Shannon, and Earth Mover’s distance). Using PSI features, we form distributionally homogeneous groups of clients via K-means++; the number of optimal clusters is chosen by a systematic silhouette-based procedure, typically yielding few clusters with modest overhead. Across six datasets (tabular, image, and text modalities), two partition protocols (Dirichlet with parameter $\alpha$ and Similarity with parameter $S$), and multiple client sizes, Clust-PSI-PFL delivers up to 18\% higher global accuracy than state-of-the-art baselines and markedly improves client fairness by a relative improvement of 37\% under severe non-IID data. These results establish PSI-guided clustering as a principled, lightweight mechanism for robust PFL under label skew. Code will be released upon acceptance.

\end{abstract}


\section{Introduction}
\label{sec:intro}
\begin{figure*}[ht]
\centering
\includegraphics[width=0.9\textwidth]{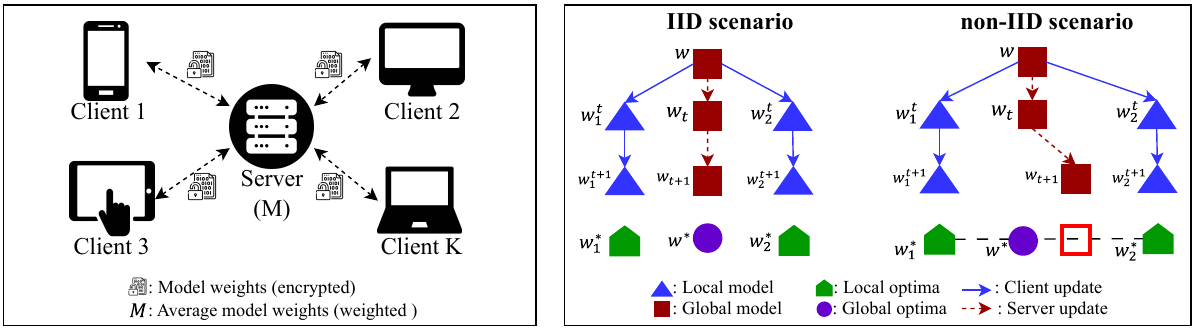}
\caption{Standard FedAvg training overview (left) and depiction of model/weight drift under non-IID scenarios (right).}
\label{fig:fl-diagram}
\end{figure*}

Federated learning (FL)~\cite{mcmahan2017communication} enables collaborative training across many clients while keeping raw data local by aggregating model updates instead of centralizing examples (so-called CL); however, non-independent and identically distributed (non-IID) data can induce biased updates and degrade global performance~\cite{g2024noniiddatafederatedlearning} (see Fig.~\ref{fig:fl-diagram}). Personalized FL (PFL) mitigates these issues by tailoring models to clients; methods are commonly grouped into (i) regularization-based approaches that keep local objectives close to the global model, (ii) selection-based schemes that prioritize or filter clients during training, and (iii) clustering-based techniques that partition clients with similar data distributions and train one model per cluster~\cite{tan2022towards}. Our approach belongs to the \emph{clustering-based} family.

This work uses the Population Stability Index (PSI), commonly employed in credit scoring to monitor distributional drift~\cite{yurdakul2018statistical,du2023proposed}, as a principled tool for client clustering in FL. PSI provides a quantitative measure of cross-client data distributional divergence, which we leverage to identify more homogeneous groups for training. We instantiate and assess this mechanism under label skew, a form of non-IID data known to most strongly degrade FL performance~\cite{g2024noniiddatafederatedlearning}. Extending the PSI-based clustering procedure to additional skew types (e.g., attribute or quantity skew) is left for future work. We deliberately start with label skew, since it is among the most practically impactful and challenging forms of non-IID data in FL. We avoid claiming generality beyond this setting and discuss extensions in Section~\ref{sec:limitations}.

\paragraph{\textbf{Motivation}}
Non-IID data poses a core challenge for FL: decentralized training inherits non-IID data that can bias local updates, slow or destabilize convergence, and degrade the aggregated global model. Detecting and mitigating non-IID data is therefore essential for fairness, robustness, and accuracy in practical deployments.

Diagnosing non-IID data in FL remains an open challenge. Surveys by Pei et al.~\cite{pei2024review} and Li et al.~\cite{li2020federated} emphasize 4 priorities: (i) lightweight, a priori diagnostics; (ii) privacy-preserving FL analysis; (iii) principled quantification of non-IID degree; and (iv) leveraging these formalizations to design optimization methods with stronger performance and, ideally, convergence guarantees.

Beyond diagnosing and quantifying, non-IID data poses a threat to client fairness: a model can achieve strong global accuracy while systematically underperforming for underrepresented clients. Therefore, we prioritize fairness alongside accuracy, measuring global accuracy and client fairness.

\paragraph{\textbf{Key insights and contributions}}
The key contributions of this work are summarized below:

\begin{enumerate}
\item We employ the PSI as a practical, low-overhead metric to quantify the level of non-IID data in FL, and analyze its properties by contrasting it with existing non-IID alternative metrics.
\item We introduce \textbf{Clust-PSI-PFL}, a clustering-based personalization framework that leverages PSI to partition clients and train per-cluster models, thereby mitigating the degradation caused by increasing non-IID data.
\item Across non-IID scenarios, Clust-PSI-PFL yields up to \textbf{18\%} boost in global test accuracy with an improvement of \textbf{37\%} in fairness over the state-of-the-art baselines.
\end{enumerate}

To the best of our knowledge, this is the first study to integrate PSI into FL as a basis for client clustering to enhance global model performance. While PSI has previously been used to guide client selection in non-IID PFL~\cite{jimenez2025psi}, we instead leverage PSI-derived data to cluster clients into distributionally homogeneous groups and train cluster-specific models.



\section{Background and Related Work}
\label{sec:related}
\paragraph{\textbf{Basics of FL}}
\label{sec:problem}

In Fig.~\ref{fig:fl-diagram} (left), the standard round-based FL pipeline shows: a server that initializes a global model and broadcasts it to \(K\) clients; each trains locally on private data and returns updates (optionally protected via multiparty secure computing (MPC)~\cite{bohler2021secure} or differential privacy (DP)~\cite{apple_privacy_2017}); the server aggregates the model's weight (e.g., FedAvg) to form a new global model and rebroadcasts, repeating until convergence.

Non-IID client data is a primary challenge in FL~\cite{g2024noniiddatafederatedlearning}. Unlike IID scenarios, non-IID client distributions (Fig.~\ref{fig:fl-diagram}, right) induce client drift: starting from the broadcast global model $w_t$, local training drives client models $w_i^{t+1}$ toward different client-specific optima $w_i^\star$, rather than toward the global optimum $w^\star$. This misalignment makes client updates point in inconsistent directions, slowing convergence and yielding suboptimal solutions. In addition, aggregating these heterogeneous updates can steer the server update away from $w^\star$, causing the global iterate (e.g., $w_{t+1}$) to move toward an undesirable region or poor local solution (highlighted in the figure as a red square), thereby further degrading accuracy.

Differences in per-client label distributions are a primary source of difficulty for FL, adversely impacting both accuracy and convergence dynamics~\cite{aggregators_FL,hsieh2020non}. To quantify the severity of non-IID data, several measures have been proposed~\cite{g2024noniiddatafederatedlearning}. Within PFL, three predominant mitigation families are widely studied: regularization-based techniques, client-selection strategies, and clustering-based solutions~\cite{tan2022towards}, as discussed in the following sections.

\paragraph{\textbf{Non-IID Data Quantification in FL}}
Prior work measures client-distribution mismatch with three common metrics. Hellinger distance (HD) estimates deviation from a balanced reference and has shown strong discrimination, particularly in cross-device FL~\cite{tan2023privacy,jimenez2024fedartml}. Jensen--Shannon distance (JSD) supports high-level comparisons and the construction of client-to-client similarity graphs to mitigate label-skew effects~\cite{ahmed2023semisupervised,xu2024fblg}. Earth Mover's/Wasserstein distance (EMD) is used to guide client scheduling and assess non-IID severity by comparing client and global label distributions~\cite{chen2022emd}.

\paragraph{\textbf{Regularization-based Baselines for Non-IID Data}}
One of the most widespread solutions to tackle the non-IID data effects is FedProx~\cite{li2020federated}. It generalizes FedAvg by adding a proximal term \(\mu\ge 0\) that controls the penalization strength. This penalizes drift and stabilizes optimization under non-IID data.

Under the FedOpt framework~\cite{reddi2020adaptive}, clients perform local stochastic gradient descent (SGD), while the server uses an adaptive update (e.g., Adagrad, Yogi, Adam) to the aggregated weights, thereby improving stability and effectiveness under non-IID data. The variants differ in moment accumulation/normalization (e.g., Yogi’s conservative second moment), offering robustness to sparse gradients and non-convexity and reducing hyperparameter sensitivity compared to FedAvg, as supported by theory and experiments.

FedAvgM~\cite{hsu2019measuring} augments FedAvg with server-side momentum, smoothing aggregated updates, and stabilizing training under non-IID data. By maintaining a moving average of the global update direction, it reduces oscillations on skewed partitions and often surpasses FedAvg. The trade-off of FedAvgM is additional hyperparameters (momentum coefficient, server learning rate) that require careful tuning.

\paragraph{\textbf{Selection-based Baselines for Non-IID Data}}

To mitigate non-IID effects, Power-of-choice (PoC)~\cite{pmlr-v151-jee-cho22a} preferentially samples high-loss clients, balancing convergence speed and bias; it achieves up to \(3\times\) faster convergence and $\sim10\%$ higher test accuracy than random selection. In contrast to PoC, our PSI-guided clustering focuses training within distributionally coherent client clusters, further improving accuracy and efficiency.

HACCS~\cite{wolfrath2022haccs} clusters clients by data-histogram similarity and, within each cluster, selects low-latency participants, preserving distributional coverage while favoring fast devices. This yields robustness to individual dropouts (as long as similar clients remain), maintains balanced representation, and accelerates convergence.

FedCLS~\cite{li2022fedcls} guides client sampling using group-level label information and Hamming distances between one-hot label vectors, preferentially selecting clients with complementary (diverse) label distributions. Compared to FedAvg with random selection, it achieves faster convergence, higher accuracy, and more stable training (well-suited to non-IID federated settings).

\paragraph{\textbf{Clustering-based Baselines for Non-IID Data}}
Clustered Federated Learning (CFL), proposed by Sattler et al.~\cite{sattler2020clustered}, tackles non-IID data by recursively clustering clients via the cosine similarity of their local updates when training stalls (small global norm, large client norm). It bipartitions to minimize cross-cluster similarity and repeats; the method is model-agnostic, FedAvg-compatible, and requires no preset number of clusters.

FedSoft, by Ruan and Joe-Wong~\cite{ruan2022fedsoft}, extends clustered FL to soft clustering, letting each client be a mixture of source distributions and jointly learning cluster models plus per-client personalized models. It uses a proximal local objective that encodes cluster models for knowledge transfer with per-round cost comparable to standard FL. The method provides convergence guarantees and outperforms hard-clustering baselines on synthetic and real datasets. 

In contrast to CFL/FedSoft, Clust-PSI-PFL clusters clients using label-histogram PSI features computed before any gradient or update is produced, and exploits PSI's per-class decomposition for interpretable client descriptors. We also select the number of clusters via a silhouette criterion, rather than CFL-style recursive splitting or FedSoft's soft-mixture optimization.

\paragraph{\textbf{Limitations of State-of-the-art Approaches}}

Progress in non-IID FL spans diagnosis (HD/JSD/EMD), stabilization (FedProx/FedOpt/FedAvgM), client selection (PoC/HACCS/FedCLS), and clustering (CFL/FedSoft). Yet gaps persist: metrics are largely descriptive and label-histogram–based (missing intra-class shifts); regularizers still bias toward a single global model and require tuning; selection via proxies can skew participation and seldom enforces within-round homogeneity; clustering is threshold-sensitive (hard) or complex and tuning-heavy (soft).

We build on and complement these advances by quantifying client-distribution dissimilarities with the PSI to form homogeneous client clusters for FL training, thereby improving performance relative to prior approaches. In our evaluation, we include all aforementioned methods as baselines and additionally report CL and FedAvg.

\section{Clust-PSI-PFL Proposed Methodology}
\label{sec:strategy}
\begin{figure*}[ht]
\centering
\includegraphics[width=0.9\textwidth]{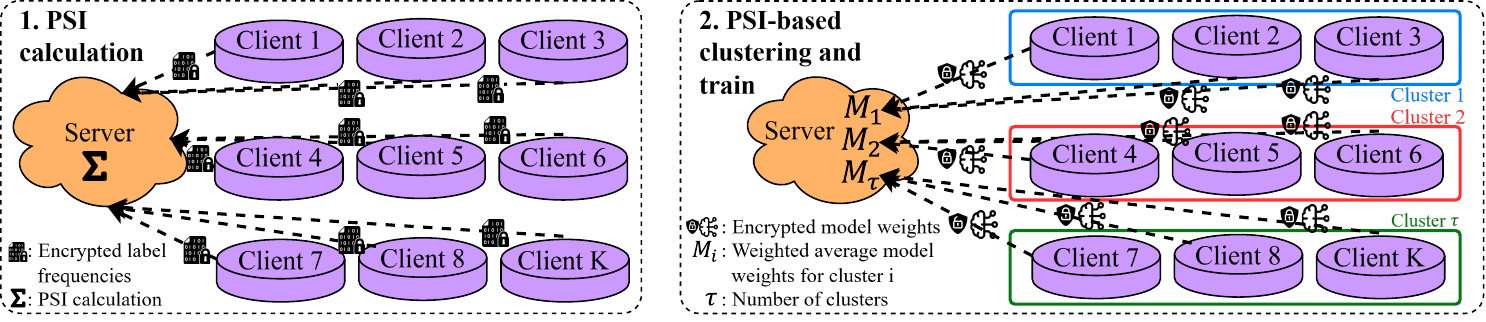}
\caption{Overview of the Clust-PSI-PFL procedure.}
\label{fig:high-level-psi-pfl}
\end{figure*}

This section presents our Clust-PSI-PFL approach; a high-level workflow is shown in Fig.~\ref{fig:high-level-psi-pfl}. Before training, every client transmits a compact label-frequency count (not raw data) to the server. We assume these per-client label counts can be shared in our threat model. When they are deemed sensitive, the same pipeline is compatible with MPC or DP noise added to histogram counts. The server aggregates these frequencies to form a reference label distribution and computes the PSI to quantify each client’s divergence from that reference. Based on the PSI metric, clients are grouped into distributionally homogeneous clusters, and a dedicated model is trained per cluster.

\subsection{Calculating PSI}
In the label-skew case, we quantify how much each client’s label distribution differs from the global one using the client-level $PSI$~\cite{siddiqi2012credit}, given in Eq.~\ref{eq:psi_formula}. This quantity is equivalent to the Jeffreys divergence~\cite{Jeffreys1946InvariantPrior}:

\begin{equation} \label{eq:psi_formula}
\begin{aligned} 
     PSI_i^L = \sum_{c=1}^{C}{\left(P(y=c) - P_i(y=c)\right)\ln\left(\frac{P(y=c)}{P_i(y=c)}\right)}
\end{aligned}
\end{equation}

\noindent Here, $P(y{=}c)$ is the server-side (global) probability mass function (pmf) for class $c$, obtained by aggregating the label counts reported by clients, while $P_i(y{=}c)$ is the corresponding pmf on client $i$; $C$ denotes the number of classes. The superscript $L$ emphasizes that we consider label skew (other skew types are left for future work).

To capture which classes contribute most to a client’s mismatch, we also use the per-class terms obtained by isolating the summands in Eq.~\ref{eq:psi_formula}:

\begin{equation}
\label{eq:psi_perclass}
\resizebox{\columnwidth}{!}{$
  PSI_{i,c}^L=\bigl(P(y{=}c)-P_i(y{=}c)\bigr)\,
  \ln\!\left(\frac{P(y{=}c)}{P_i(y{=}c)}\right),\quad c=1,\ldots,C
$}
\end{equation}

The vector $[\,PSI_i^L,\, PSI_{i,1}^L,\,\ldots,\,PSI_{i,C}^L\,]$ is then used as a client descriptor for clustering.

Finally, we summarize federation-level label non-IID data with the sample-size weighted average $WPSI^L$:

\begin{equation} \label{eq:wpsi_formula}
\begin{aligned} 
     WPSI^L = \sum_{i=1}^{K}{\frac{n_i}{N}PSI_i^L}
\end{aligned}
\end{equation}

\noindent where $K$ is the number of clients, $n_i$ is the local sample count, and $N=\sum_{i=1}^{K} n_i$. Smaller $PSI_i^L$ (and $WPSI^L$) indicate more homogeneous clients, whereas larger values reflect stronger non-IID label skew.

\subsection{PSI-based Cluster Creation and Training}
Upon computing the client PSI at the server, we partition clients into clusters according to their PSI signatures. Since PSI quantifies distributional divergence, grouping clients with similar PSI profiles forms more homogeneous cohorts. Aggregating updates within such cohorts reduces cross-distribution mixing, mitigates the weight–update mismatch induced by non-IID data, and improves convergence stability and speed.

Clients are clustered with K-means++, selected for its robust seeding and broad adoption~\cite{ArthurVassilvitskii2007KMeansPP}. Each client is encoded as a feature vector that concatenates the overall PSI and the per-class PSI values (Eqs.~\ref {eq:psi_formula} and~\ref{eq:psi_perclass}). These features are standardized to have a mean of zero and a variance of one. This step is lightweight, adding minimal server-side overhead, since its cost scales mainly with the (typically small) number of classes $C$.



The cluster count $\tau$ is a key hyperparameter in Clust-PSI-PFL. To avoid ad hoc choices, we select $\tau$ using the PSI-based algorithm for selecting the number of clusters, which is aligned with the statistical characteristics of the client population. Given the resulting assignments $z^{(\tau)}$, we partition clients into $\tau$ clusters and train a FedAvg-type FL model for each cluster; the efficacy of this design is validated empirically in Section~\ref{sec:experiments}.

\noindent\emph{Silhouette score~\cite{rousseeuw1987silhouettes}.}
To quantify the quality of a candidate clustering with $j$ clusters, we use the (average) silhouette score.
For each client feature vector $x_i$, let $a(i)$ denote the average distance from $x_i$ to all other points in its assigned cluster (intra-cluster cohesion), and let $b(i)$ denote the minimum, over all other clusters, of the average distance from $x_i$ to the points in that other cluster (nearest-cluster separation).
The silhouette coefficient of $x_i$ is defined as $\mathrm{sil}(i) = \frac{b(i)-a(i)}{\max\{a(i),b(i)\}}$, which lies in $[-1,1]$, with larger values indicating better separation and cohesion.
We report the mean silhouette over all clients as $s(j)$ and select $\tau = \arg\max_j s(j)$.

\begin{algorithm}[t]
\caption*{\textbf{PSI-based algorithm for selecting the number of clusters:} Selecting $\tau$ via PSI and the silhouette score.}
\label{alg:psi-threshold}
\begin{algorithmic}[1]
\STATE \textbf{Input:} Clients $\mathcal{I}=\{1,\ldots,K\}$; for each client $i$: overall $PSI_i^L$ and per-class $PSI_{i,c}^L$ for $c=1,\ldots,C$ (number of classes)
\STATE \textbf{Output:} Selected number of clusters $\tau$ and cluster assignment $z^{(\tau)}$
\STATE Build feature matrix $X \in \mathbb{R}^{K \times (C+1)}$, with row $x_i = [\,PSI_i^L, PSI_{i,1}^L, \ldots, PSI_{i,C}^L\,]$
\STATE Standardize features of $X$ to zero mean and unit variance
\STATE $best\_score \gets -\infty$; $\tau \gets 2$; $z^{(\tau)} \gets \varnothing$
\FOR{$j = 2$ to $K-1$}
  \STATE Run K-means with K-means++ initialization on $X$ to obtain candidate clusters $z^{(j)}$
  \STATE Compute average silhouette score $s(j)$ on $X$ using the candidate clusters $z^{(j)}$
  \IF{$s(j) > best\_score$}
    \STATE $best\_score \gets s(j)$; $\tau \gets j$; $z^{(\tau)} \gets z^{(j)}$
  \ENDIF
\ENDFOR
\STATE \textbf{return} $\tau$, $z^{(\tau)}$
\end{algorithmic}
\end{algorithm}

We provide a concise analysis of the computational complexity of the Clust-PSI-PFL clustering phase. The cost arises mainly from (i) computing PSI for each client and (ii) selecting the number of clusters via K-means++ with the silhouette score. Computing PSI over \(K\) clients and \(C\) classes requires \(\mathcal{O}(K\,C)\) operations. Let \(I\) be the average number of K-means++ iterations, and let the search examine \(K-2\) candidate cluster counts \(j\in\{2,\ldots,K-1\}\). A single K-means++ run with \(j\) clusters costs \(\mathcal{O}(I\,K\,j\,(C+1))\), while computing the silhouette costs \(\mathcal{O}(K^{2}(C+1))\). Summed over the \(K-2\) candidates, the model-selection stage costs \(\mathcal{O}\!\big((C+1)\big(I\,K\sum_{j=2}^{K-1} j + (K-2)\,K^{2}\big)\big)=\mathcal{O}\!\big((C+1)\big(I\,K\,((K-1)K/2 - 1) + (K-2)\,K^{2}\big)\big)\). Therefore, the total clustering complexity is \(\mathcal{O}\!\big(K\,C + (C+1)\big(I\,K\,((K-1)K/2 - 1) + (K-2)\,K^{2}\big)\big)\). 
In practice, the number of clusters explored is small, and this selection is performed only once before FL training, so the added cost is amortized over all communication rounds. In very large-scale deployments, silhouette scores can be approximated on a subsample of clients with negligible impact on the selected number of clusters. Subsequent training proceeds as in FedAvg, except that aggregation is performed independently for each of the \(\tau\) clusters.

\subsection{PSI vs. Alternative Non-IID Metrics}

The weighted PSI (\(WPSI^L\)) is well-suited metric to quantify non-IID in FL. We benchmark it against HD, JSD, and EMD~\cite{jimenez2024fedartml}, standard divergences for comparing distributions (see Section~\ref{sec:related}). Unlike these baselines, \(WPSI^L\) delivers fine-grained, client-level diagnostics with minimal overhead and, thanks to PSI’s classwise decomposition, enables per-class attribution to pinpoint labels driving divergence, capabilities HD/JSD/EMD generally lack without extra modifications. Thus, \(WPSI^L\) is an interpretable and efficient primary metric for non-IID assessment.

\begin{figure}[t]
\centering
\subfloat[Dirichlet]{
  \includegraphics[width=0.96\columnwidth]{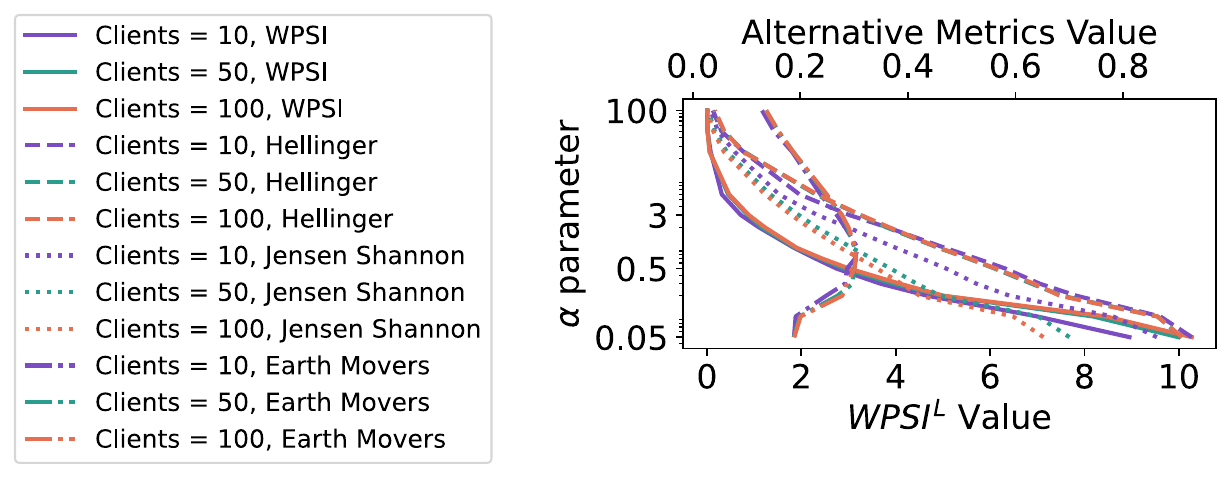}
  \label{fig:wpsi_dirichlet}
}
\hfil
\subfloat[Similarity]{
  \includegraphics[width=0.96\columnwidth]{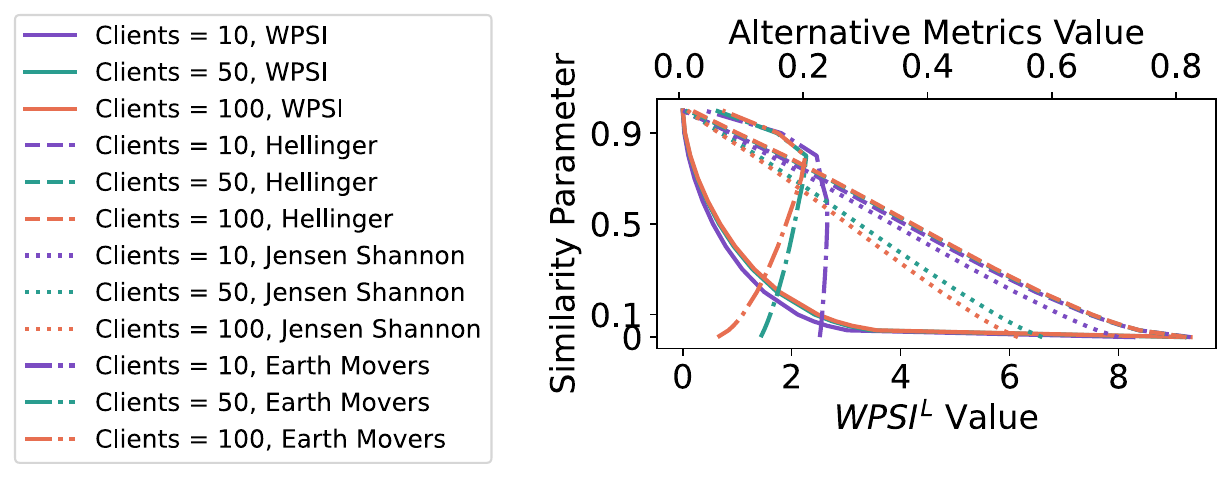} 
  \label{fig:wpsi_similarity}
}
\caption{Mapping between $WPSI^L$ and non-IID level under the Dirichlet and Similarity partition protocols.}
\label{fig:wpsi_noniidness_both}
\end{figure}

To evaluate the practical utility of $WPSI^L$, we conducted a broad empirical study spanning multiple datasets, levels of non-IID data, partitioning schemes, and random seeds (see Section~\ref{sec:experiments}). Centralized datasets were partitioned using the Dirichlet~\cite{jimenez2024fedartml} and Similarity~\cite{wang2023distribution} partition protocols, after which we computed $WPSI^L$, HD, JSD, and EMD. 
Fig.~\ref{fig:wpsi_noniidness_both} shows the relationship between $WPSI^L$ and the degree of non-IID data, controlled by the Dirichlet parameter $\alpha$ and the Similarity parameter $S$, with the corresponding values of alternative metrics (HD, JSD, EMD) reported on the secondary axis. Our findings are as follows: (i) $WPSI^L$ decreases with increasing $\alpha$ and $S$ (i.e., as data become more IID), following a pronounced decay under both protocols; (ii) JSD and HD vary more smoothly and approximately linearly with $WPSI^L$ across both schemes; and (iii) EMD is less reliable as a non-IID quantifier due to non-monotonic behavior in some settings, potentially yielding identical values at distinct non-IID levels.


\begin{figure}[t]
\centering
\includegraphics[width=0.95\columnwidth]{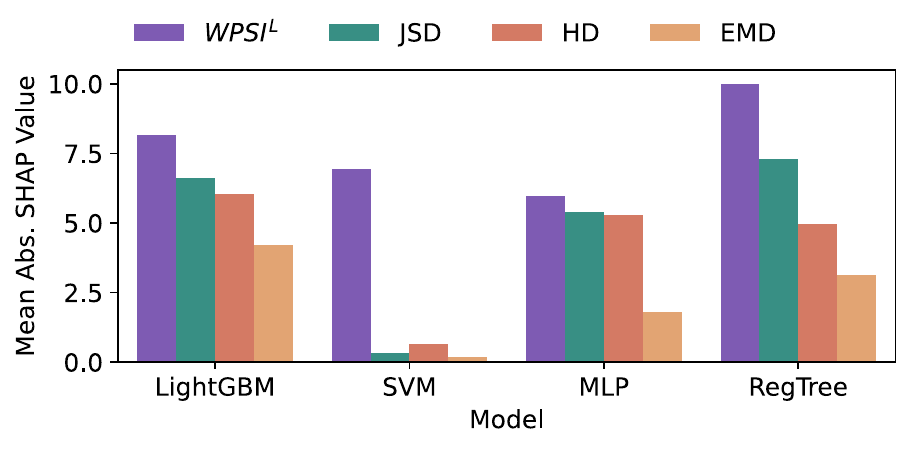}\\[0.6em]
\includegraphics[width=0.95\columnwidth]{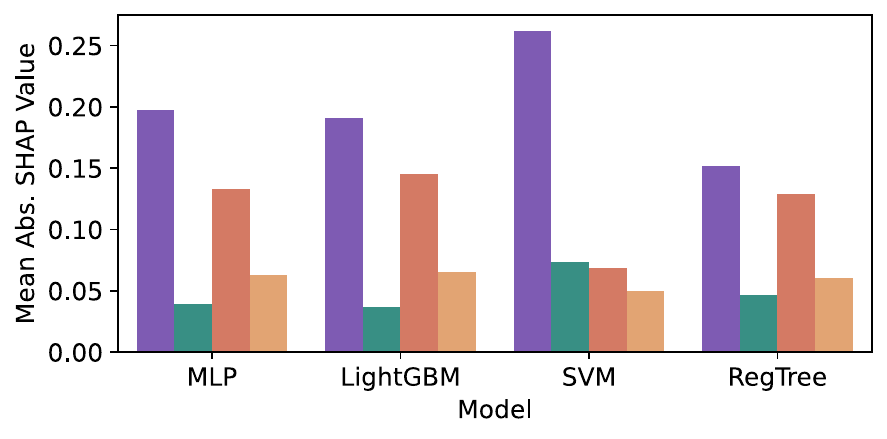}
\caption{Feature importances for predicting non-IID level under Dirichlet (top) and Similarity (bottom) partition protocols.}
\label{fig:wpsi_feat_imp_all_CLI_LS_all_Alp_vert}
\end{figure}


Using the computed metrics as features, we trained LightGBM, Support Vector Machine (SVM), Multilayer Perceptron (MLP), and a Regression Tree to predict the Dirichlet parameter $\alpha$ and the Similarity parameter $S$. Feature-importance scores from these models (Fig.~\ref{fig:wpsi_feat_imp_all_CLI_LS_all_Alp_vert}) consistently ranked $WPSI^L$ highest, indicating it is the most informative measure of non-IID structure.

\vspace{0.5em}
\noindent \vspace{0.1em} \mybox{gray}{\observation{$WPSI^L$ consistently appears as the strongest predictive metric for the degree of non-IID data.}}

\section{Experiments and Results}
\label{sec:experiments}
In this section, we describe the simulation environment, datasets, and evaluation protocol to enable reproducibility, and subsequently present the main results of our analyses.

We investigate the following three empirical questions (EQs), which probe core properties of Clust-PSI-PFL and its comparison to state-of-the-art baselines:

\begin{itemize}
\item \textbf{EQ1:} How does the silhouette-based procedure for selecting the optimal number of clusters behave across different client configurations?
\item \textbf{EQ2:} Does Clust-PSI-PFL consistently surpass competing baselines in global accuracy under varying degrees of non-IID data?
\item \textbf{EQ3:} To what extent does Clust-PSI-PFL improve client fairness (see Section~\ref{subsec:metrics}) relative to state-of-the-art baselines, especially in highly non-IID data partitions?
\end{itemize}

Each EQ is addressed by a concise observation presented in a gray box.

\subsection{Experimental Design}

\paragraph{\textbf{Testbed}}
All experiments ran on a single Linux machine (Beastbianv2, kernel 6.1.0-20-amd64) with an AMD Threadripper PRO 5995WX CPU (64 cores), 500\,GB RAM, and an NVIDIA RTX A6000 GPU (48\,GB). The codebase was implemented in Python~3.10 and relied on Flower~\cite{beutel2020flower} and FedLab~\cite{zeng2023fedlab} for the FL baselines.

\paragraph{\textbf{Data and Partition Protocols}}
We assess Clust-PSI-PFL across a broad suite of datasets spanning tabular, image, and text modalities, using diverse models and partition protocols. The specifications of the centralized datasets are summarized in Table~\ref{tab:characteristics_datasets}. These datasets were selected for their widespread use in prior studies on non-IID phenomena in FL.

We employ two widely used partitioning protocols to simulate varying client data distributions. Dirichlet~\cite{jimenez2024fedartml} controls non-IID data with a single parameter $\alpha$: a smaller $\alpha$ generates more non-IID client partitions. In addition, we use the Similarity partition protocol~\cite{wang2023distribution}, which is governed by a single parameter $S\!\in\![0,1]$: first, allocate $S*100\%$ of the dataset uniformly at random to clients (IID component); then sort the remaining $(100{-}S*100)\%$ by label and distribute it evenly across clients (label-skewed component). Larger $S$ produces more IID partitions, while smaller $S$ induces pathological non-IID data.

\begin{table}[t]
    \centering
    \caption{Summary of datasets properties}
    \resizebox{\columnwidth}{!}{
    \begin{tabular}{||ccccc||}
    \hline
    \textbf{Dataset} & \textbf{Modality} & \makecell{\textbf{Task}} & \textbf{\makecell{\# classes}} & \textbf{\makecell{\# examples}} \\
    \hline \hline
    \makecell{ACSIncome~\cite{ding2021retiring}} & Tabular & \makecell{Classify high/low salary}  & 2 & 1,664,500 \\
    \hline    
    Serengeti~\cite{swanson2015snapshot} & Tabular & \makecell{Detect animal type}  & 13 & 286,586 \\
    \hline
    FMNIST~\cite{xiao2017fashion} & Image & \makecell{Detect clothing type} & 10 & 70,000 \\
    \hline
    CIFAR10~\cite{krizhevsky2009learning} & Image & \makecell{Classify objects} & 10 & 60,000 \\
    \hline
    \makecell{Sent140~\cite{go2009twitter}} & Text & \makecell{Sentiment analysis} & 3 & 1,600,000 \\
    \hline
    \makecell{Amazon reviews~\cite{zhang2015character}} & Text & \makecell{Product classification} & 6 & 50,000 \\    
    \hline
    \end{tabular}}
    \label{tab:characteristics_datasets}
\end{table}

\paragraph{\textbf{Selection of $\alpha$ and $S$ Values}}
The $\alpha$ values were chosen to cover the entire spectrum of non-IID settings, measured by $WPSI^L$ ranging from 0 (IID) to large values (extreme non-IID). We tested eleven $\alpha$ values within this range, but, for brevity, present results for $\alpha = \{50,0.7,0.3,0.2,0.09,0.05\}$, as they are representative and consistent with the overall findings. Note that the effect of the Dirichlet $\alpha$ parameter is dataset-dependent: the attainable degree of non-IID data varies with the number of classes. In datasets with few classes, very small values (e.g., $\alpha<0.3$) are infeasible. The $S$ values for the Similarity partition protocol were selected to span the full non-IID data range, from $S=0$ (maximally non-IID) to $S=1$ (IID). We evaluated eleven $S$ settings across this interval, but, for brevity, report results for $S \in \{1, 0.03, 0\}$, which are representative of the overall trends.


\paragraph{\textbf{Models}} For ACSIncome, we train a shallow classifier consisting of a single fully connected layer with a softmax/sigmoid output (i.e., a logistic-regression baseline). 
For Serengeti, we use a feed-forward network with three hidden layers of 500 neurons each and ReLU non-linearities, ending with a softmax classification head. 
For FMNIST, we rely on a lightweight CNN made of three convolutional stages with 8, 16, and 32 feature maps, respectively; the convolutional stack is followed by max-pooling and a dense layer with 2048 ReLU units before the final classifier. 
For CIFAR-10, we adopt a deeper convolutional architecture composed of three feature-extraction blocks. In each block, two $3\times3$ convolutions (ReLU, same padding) are applied and regularized with batch normalization, after which a $2\times2$ max-pooling operation reduces spatial resolution. We further apply dropout after each block, increasing the rate from 0.2 to 0.3 and 0.4 in the successive blocks, while using 32, 64, and 128 filters in the first, second, and third blocks, respectively. The final activation maps are flattened and fed to a 128-unit fully connected ReLU layer with batch normalization and dropout ($p{=}0.5$), and a softmax output over the $C$ classes (input size $H\times W\times3$).

\begin{table}[t]
    \centering
    \caption{Hyperparameters description and selection}
    \resizebox{\columnwidth}{!}{
    \begin{tabular}{||cccc||}
    \hline
    \textbf{Baseline} & \textbf{Hyperparameter} & \textbf{Grid range} & \textbf{Best Value} \\ \hline
    \hline
    All & Learning rate & \{0.001,0.01,0.1\} & 0.001\\
    \hline
    All & Batch size & \{16,32,64\} & 32\\
    \hline
    All & Optimizer & \{Adam, SGD\} & Adam\\
    \hline    
    FedProx & $\mu$ & \{0.001,0.01,0.1\} & 0.01\\
    \hline
    FedAvgM & Momentum & \{0.7,0.9,0.99\} & 0.7\\
    \hline
    FedAdagrad & $\tau$ & \{0.01,0.1\} & 0.1\\
    \hline
    FedAdagrad & $\eta$ & \{0.1,0.3162\} & 0.3162\\
    \hline
    FedAdagrad & $\eta_l$ & \{0.01,1\} & 1\\
    \hline
    FedYogi & $\tau$ & \{0.001,0.01,0.1\} & 0.01\\
    \hline
    FedYogi & $\eta$ & \{0.001,0.01\} & 0.001\\
    \hline
    FedYogi & $\eta_l$ & \{0.01,1\} & 1\\
    \hline
    FedYogi & $\beta_1$ & \{0.9,0.99\} & 0.9\\
    \hline
    FedYogi & $\beta_2$ & \{0.9,0.99\} & 0.99\\
    \hline
    FedAdam & $\tau$ & \{0.001,0.01,0.1\} & 0.001\\
    \hline
    FedAdam & $\eta$ & \{0.001,0.01\} & 0.001\\
    \hline
    FedAdam & $\eta_l$ & \{0.01,1\} & 1\\
    \hline
    FedAdam & $\beta_1$ & \{0.9,0.99\} & 0.9\\
    \hline
    FedAdam & $\beta_2$ & \{0.9,0.99\} & 0.99\\
    \hline
    PoC & d & \{6,8,10,16\} & 8\\
    \hline
    HACCS & $\rho$ & \{0.1,0.5,0.95\} & 0.95\\
    \hline
    FedCLS & $\tau$ & \{0.1,0.5,0.95\} & 0.1\\
    \hline
    CFL & $\epsilon_1$ & \{0.00001,0.0001,0.001\} & 0.00001\\
    \hline
    CFL & $\epsilon_2$ & \{0.001, 0.01, 0.1\} & 0.1\\
    \hline
    CFL & $\gamma$ & \{0.1,0.5,0.9\} & 0.5\\
    \hline
    FedSoft & $n$ & \{5,10,50\} & 5\\
    \hline
    Clust-PSI-PFL (ours) & $\tau$ & \{From 2 to $K-1$\} & Optimal $\tau$\\
    \hline
    \end{tabular}}
    \label{tab:hypeparameters}
\end{table}

For Sent140, we employ a compact sequence model consisting of an embedding layer, dropout ($p{=}0.5$), a 10-unit LSTM (dropout/recurrent dropout $0.2$), and a softmax classifier. For Amazon Reviews, we use a small embedding-based DNN (embedding $\rightarrow$ flatten $\rightarrow$ 8-unit ReLU layer $\rightarrow$ softmax). All results are averaged over five independent partitions (five seeds). With standard deviations mostly below 0.02 (see Table~\ref{tab:global_dirichlet_similarity_by_dataset} and~\ref{tab:local_dirichlet_similarity_by_dataset}), additional seeds add little value; five were enough to obtain stable estimates and preserve method rankings while avoiding unnecessary computation.

The candidate hyperparameter grids and the corresponding optimal settings for each baseline are summarized in Table~\ref{tab:hypeparameters}. The search ranges were chosen in accordance with recommendations from the original baseline papers. Final values were selected via an exhaustive grid search over all parameter combinations.

We train for \(T= 40\) communication rounds; at each round, a fraction \(q=0.5\) of the \(K\) clients is sampled uniformly without replacement, and each selected client runs \(E=5\) local epochs. We use official test sets when available; otherwise, we adopt a random train/test split of \(80\%/20\%\).

\paragraph{\textbf{Metrics}}\label{subsec:metrics}

This section summarizes the evaluation metrics used in our experiments.

\emph{Accuracy.} We report the proportion of correctly classified samples; higher values correspond to better predictive performance. Since FL data are distributed, we consider both per-client and federation-level accuracy.

\emph{Local accuracy.} For each client $k$, we compute:
$A_k=\frac{CC_k}{n_k}$ 
where $CC_k$ is the number of correct predictions on client $k$ and $n_k$ is its sample count.

\emph{Global accuracy.} We obtain a single federation score by averaging local accuracies weighted by client data sizes:
$A_{global} = \frac{\sum_{k=1}^{K} n_k A_k}{\sum_{k=1}^{K} n_k}$ with $K$ the total number of participating clients and $n_k$ the number of data samples on client $k$.

\emph{Client fairness.} To characterize performance disparities across clients, we measure each client’s deviation from perfect accuracy (1.0) and summarize it via the mean distance
$AD=\frac{1}{K} \sum_{i=1}^K (|A_i-1.0|)$
and its dispersion
$SDAD=\sqrt{\frac{1}{K} \sum_{i=1}^K (|A_i-1.0| - AD)^2}$.
Lower $AD$ and $SDAD$ indicate, respectively, higher overall fairness and smaller variability in client outcomes.

\subsection{Clust-PSI-PFL performance evaluation}

This subsection evaluates the global and local performance of Clust-PSI-PFL in comparison to the baseline methods.

\paragraph{Number of Clusters Behavior}

In this subsection, we answer EQ1 by examining the behavior of the number of clusters ($\tau$) on PSI-based clusters determined via the systematic approach described in Section~\ref{sec:strategy}.

\begin{figure}[ht]
\centering
\subfloat[Dirichlet ($\alpha=0.3$)]{
  \includegraphics[width=1.0\columnwidth]{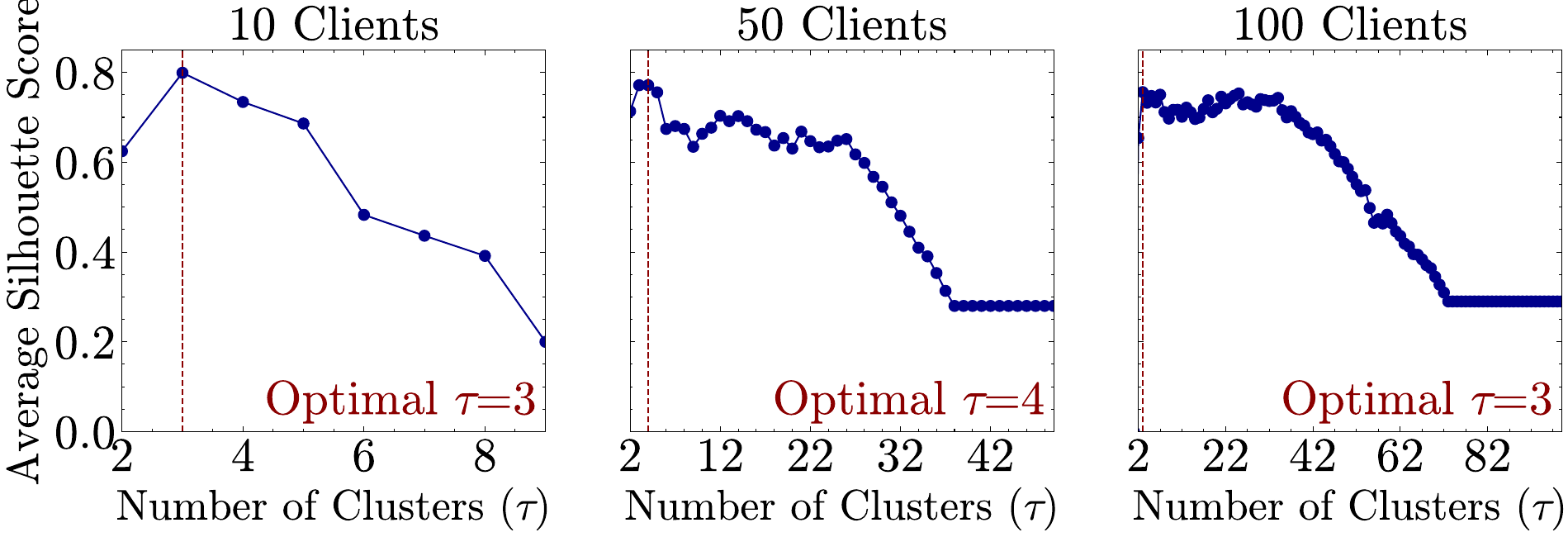}
  \label{fig:acsincome_silhouette_dirichlet_all_CLI_LS_all_Alp}
}
\hfil
\subfloat[Similarity ($S$=0)]{
  \includegraphics[width=1.0\columnwidth]{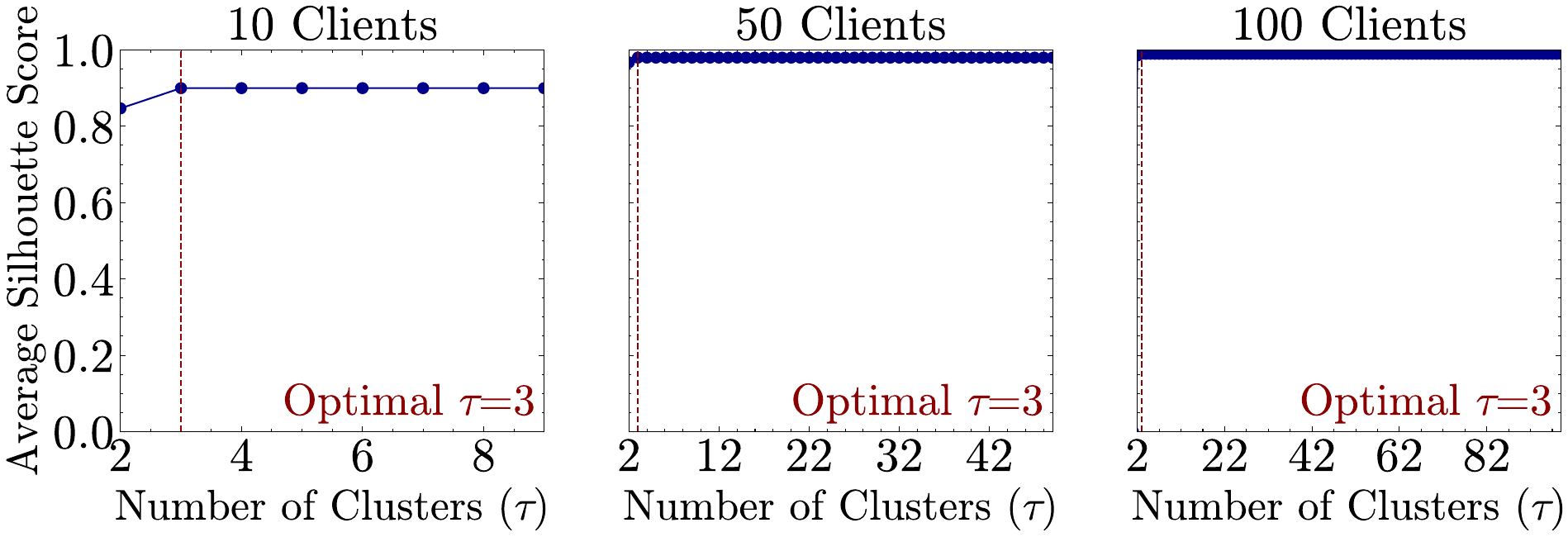} 
  \label{fig:acsincome_silhouette_similarity_all_CLI_LS_all_Alp}
}
\caption{Silhouette score for K-means++ changing number of clusters ($\tau$) for the Dirichlet and Similarity partition protocols and the ACSIncome dataset.}
\label{fig:silhouette_both}
\end{figure}

\begin{figure}[ht]
\centering
\subfloat[Dirichlet ($\alpha=0.3$)]{
  \includegraphics[width=\columnwidth]{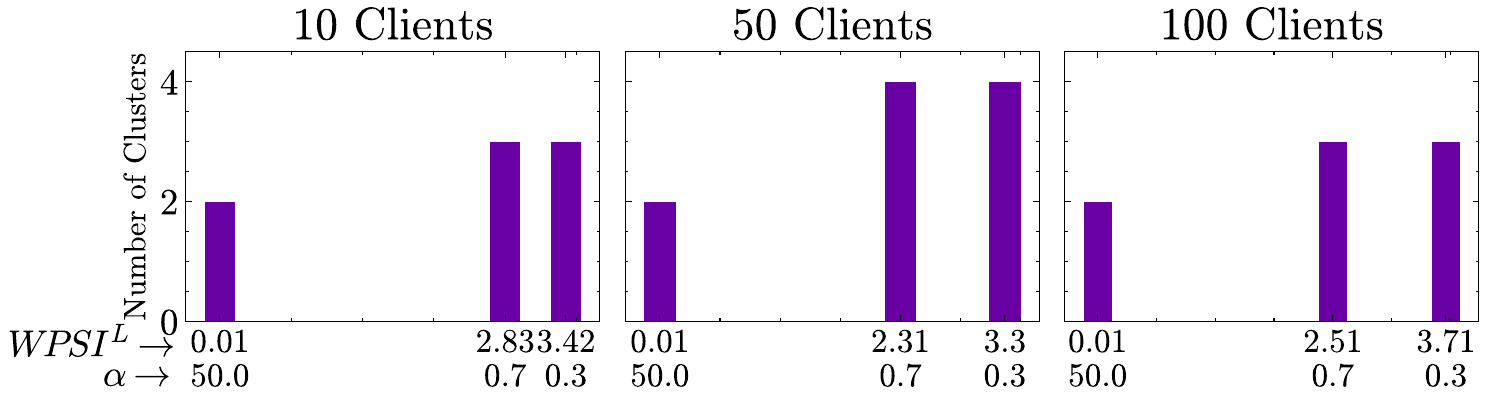}
  \label{fig:acsincome_num_clust_dirichlet_all_CLI_LS_all_Alp}
}
\hfil
\subfloat[Similarity ($S$=0)]{
  \includegraphics[width=\columnwidth]{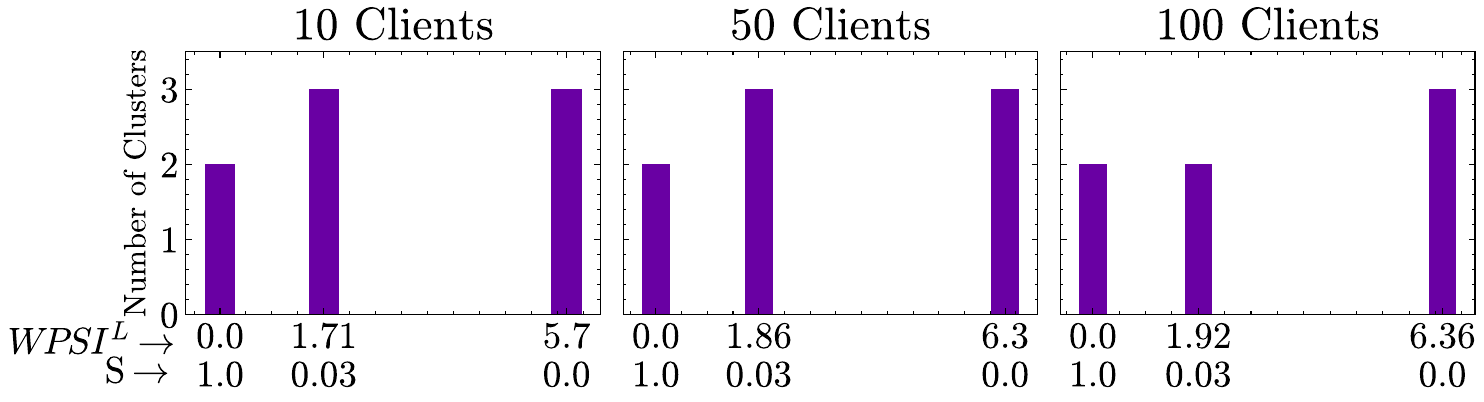} 
  \label{fig:acsincome_num_clust_similarity_all_CLI_LS_all_Alp}
}
\caption{Number of clusters ($\tau$) for the Dirichlet and Similarity partition protocols and the ACSIncome dataset.}
\label{fig:num_clusters_both}
\end{figure}

\begin{figure*}[ht]
\centering
\subfloat[Dirichlet]{
  \includegraphics[width=0.8\textwidth]{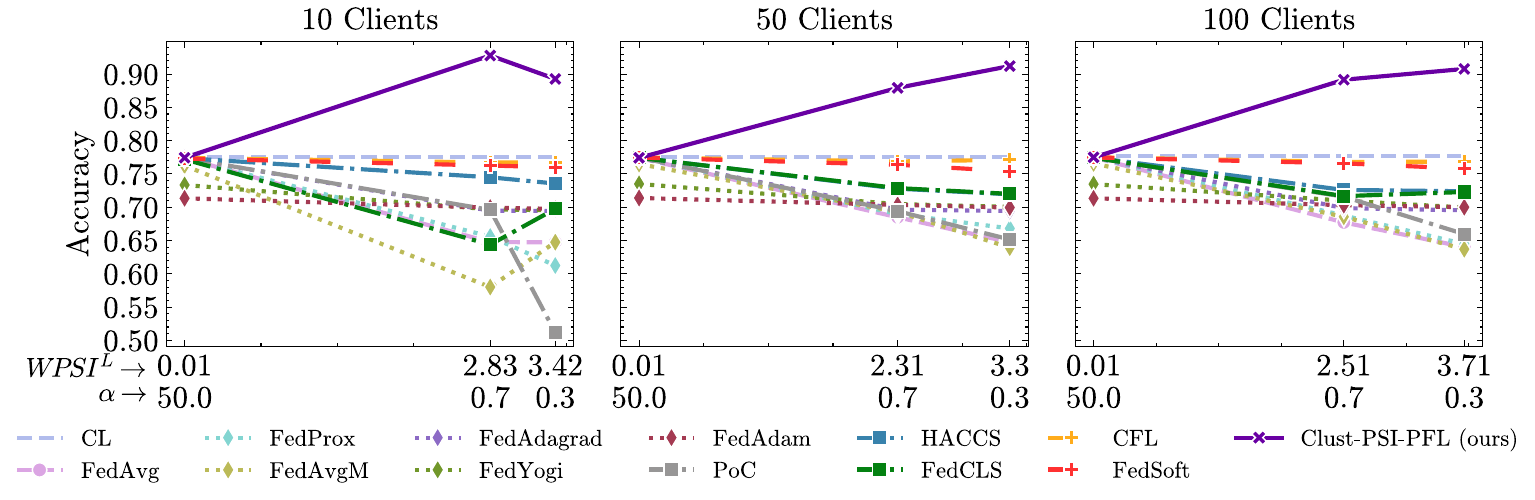}
  \label{fig:acsincome_global_acc_dirichlet_all_CLI_LS_all_Alp}
}
\hfil
\subfloat[Similarity]{
  \includegraphics[width=0.8\textwidth]{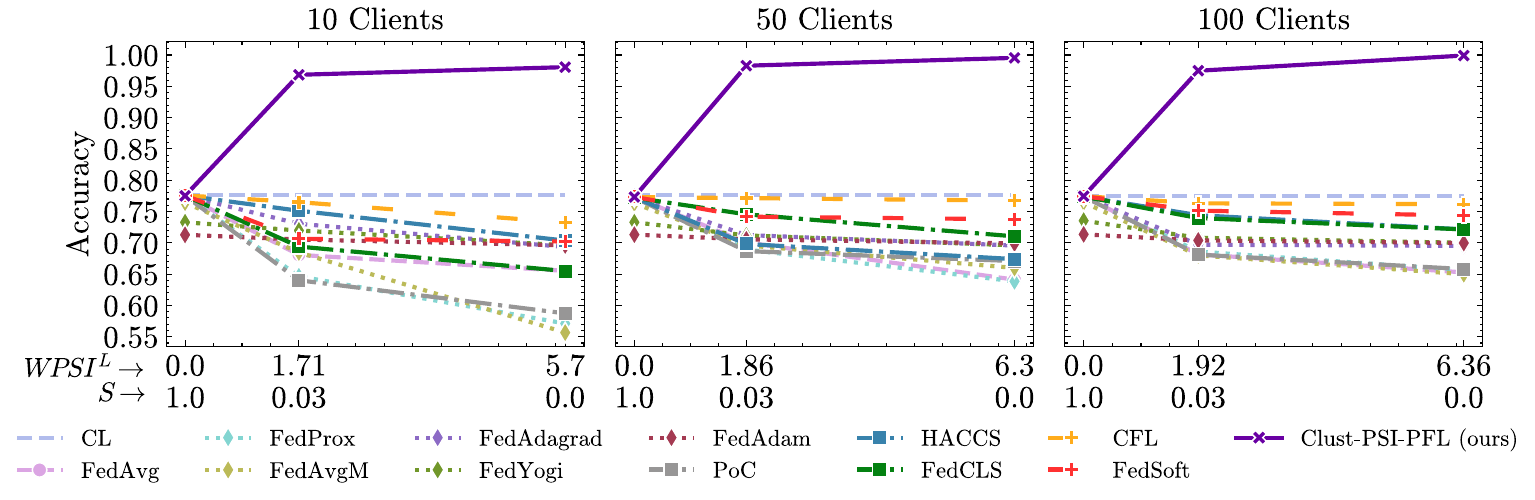}
  \label{fig:acsincome_global_acc_similarity_all_CLI_LS_all_Alp}
}
\caption{Global test accuracy for all baselines and ACSIncome dataset.}
\label{fig:global_acc_both}
\end{figure*}

Fig.~\ref{fig:silhouette_both} illustrates the silhouette score against the number of clusters for the ACSIncome dataset, changing the number of clients participating in the FL. Due to space constraints, we present only the ACSIncome dataset since its behavior closely mirrors that of the remaining datasets and non-IID data configurations.

\begin{figure*}[ht]
\centering
\subfloat[Dirichlet]{
  \includegraphics[width=0.8\textwidth]{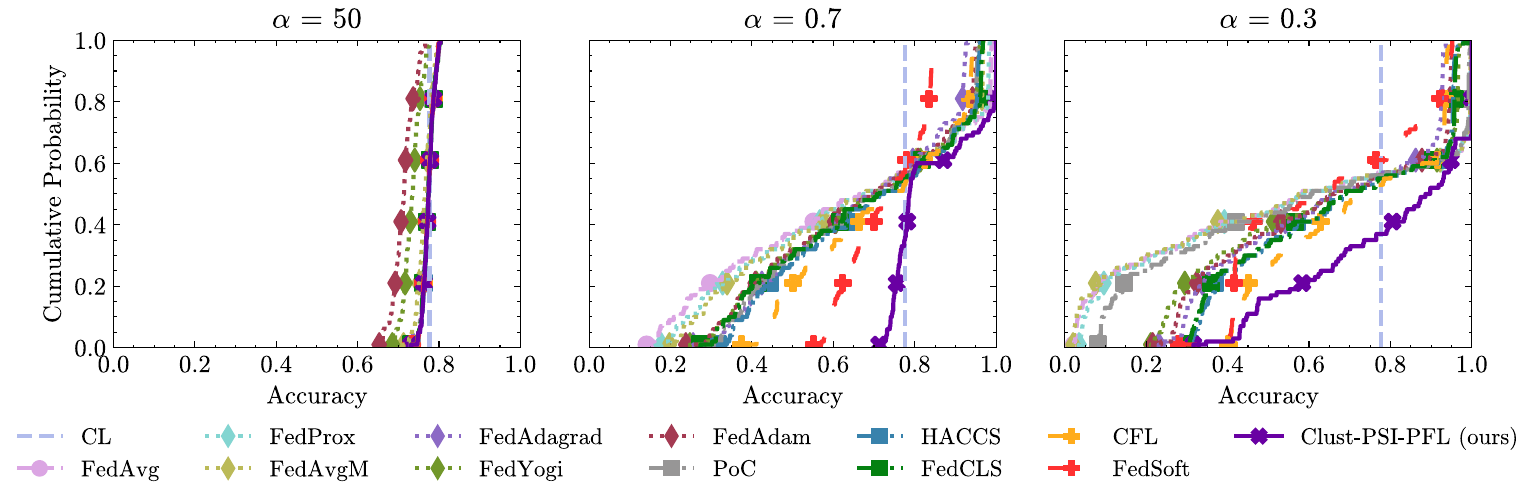}
  \label{fig:acsincome_local_acc_dirichlet_100_CLI_LS_all_Alp}
}
\hfil
\subfloat[Similarity]{
  \includegraphics[width=0.8\textwidth]{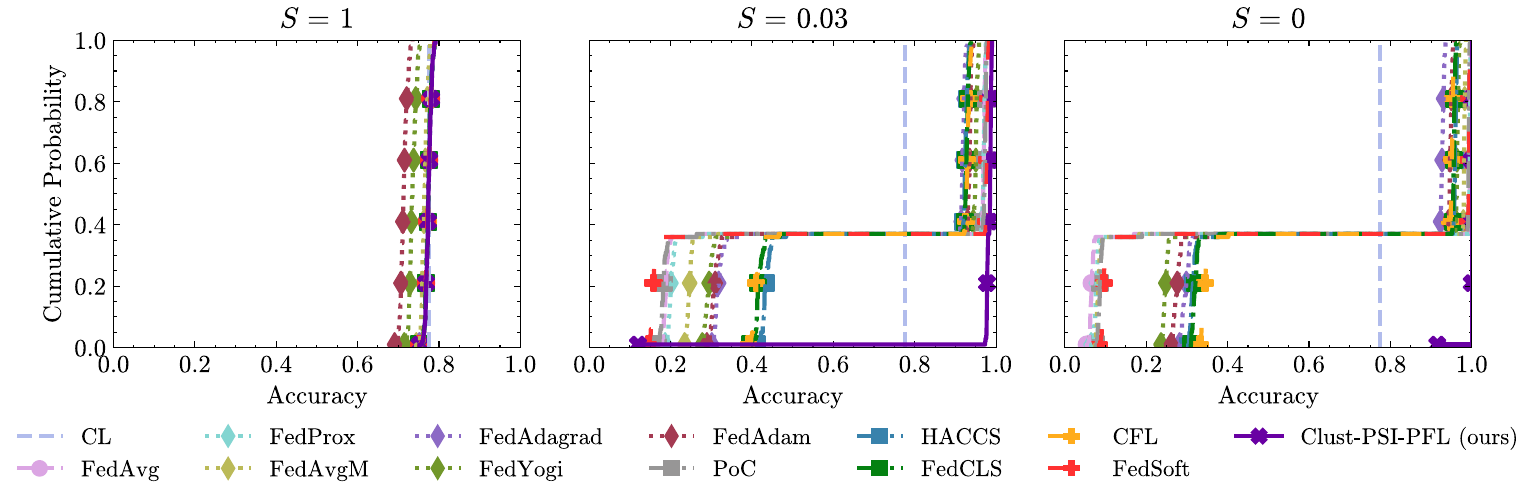}
  \label{fig:acsincome_local_acc_similarity_100_CLI_LS_all_S}
}
\caption{Local test accuracy ECDF for all baselines, ACSIncome dataset, and $K=100$ clients.}
\label{fig:local_acc_both}
\end{figure*}

Under Dirichlet (\(\alpha=0.3\)), the silhouette peaks at small \(\tau\): \(\tau=3\) for \(K=10\), \(\tau=4\) for \(K=50\), and \(\tau=3\) for \(K=100\), indicating a compact \(3\!-\!4\) cluster structure with weaker separation as \(K\) grows. Under Similarity (\(S=0\)), the curve is high and nearly flat; applying our rule (choose the smallest \(\tau\) at the maximum) yields \(\tau=3\) for all \(K\). Overall, PSI features reveal clear clusters selecting a small, stable \(\tau\), especially under pathological label skew.

Fig.~\ref{fig:num_clusters_both} reports the number of clusters \(\tau\) selected by the PSI-based algorithm for selecting the number of clusters for ACSIncome as the number of clients \(K\) and non-IID data levels vary. Under Dirichlet (\(\alpha=0.3\)), \(\tau\) remains small, typically \(3\)–\(4\), with a mild fluctuation (\(\tau=3\) for \(K=10\) and \(100\), \(\tau=4\) for \(K=50\)). Under Similarity (\(S=0\)), the selection is even more stable, yielding \(\tau=3\) across all \(K\). Thus, despite changes in client population and non-IID data (as reflected by \(WPSI^L\)), PSI-based clustering consistently uncovers a compact latent structure, keeping the number of cluster models low and the training overhead modest.

\vspace{0.5em}
\noindent \vspace{0.1em} \mybox{gray}{\observation{Across diverse client counts and partition protocols, the PSI-based algorithm for selecting the number of clusters consistently favors small $\tau$, indicating a compact structure and enabling low-overhead training.}}
\vspace{0.5em}

\paragraph{\textbf{Global-level Test Clust-PSI-PFL Behavior}}
At the global level, the server aggregates client updates (within each cluster) to produce a cluster-level model intended to generalize to the full population. This system-wide view evaluates accuracy, convergence speed, and robustness to non-IID data skew; the resulting global metrics summarize the end-to-end effectiveness of the FL pipeline across all participating clients.

To address EQ2, we analyze Fig.~\ref{fig:global_acc_both}, which reports the mean test global accuracy of Clust-PSI-PFL (purple) and all baselines as a function of the number of clients, the Dirichlet parameter $\alpha$, and the Similarity parameter $S$, together with the corresponding $WPSI^L$. Across settings, Clust-PSI-PFL consistently surpasses competing baselines, including highly non-IID data scenarios, achieving up to 18\% relative gains over strong contenders such as CFL and FedSoft. Moreover, its variability is low (standard deviation $\sim 0.01$), indicating stable training. Performance generally improves as the client population grows (e.g., $K=50$ or $K=100$), where inter-client distributional differences become more pronounced and clustering is more effective.

\begin{table*}[htbp]
\centering
\caption{Global test accuracy (mean $\pm$ standard deviation) for baselines under both partition protocols on all the datasets for $K=100$. Each simulation was run for 5 seeds (trials). ``---'' denotes infeasible partitions. Best performing shown in bold.}
\resizebox{\textwidth}{!}{
\begin{tabular}{||c|l|cccccc|ccc||}
\hline
& & \multicolumn{6}{c|}{\textbf{Dirichlet}} & \multicolumn{3}{c||}{\textbf{Similarity}} \\ \cline{3-11}
\textbf{Dataset} & \textbf{Baseline} & $\alpha=50$ & $\alpha=0.7$ & $\alpha=0.3$ & $\alpha=0.2$ & $\alpha=0.09$ & $\alpha=0.05$ & $S=1$ & $S=0.03$ & $S=0$ \\ \hline\hline

\multirow{13}{*}{ACSIncome}
 & CL                     & \multicolumn{9}{c||}{0.76 $\pm$ 0.02} \\ \cline{2-11}
 & FedAvg                 & 0.77 $\pm$ 0.01 & 0.67 $\pm$ 0.02 & 0.64 $\pm$ 0.01 & --- & --- & --- & 0.77 $\pm$ 0.01 & 0.68 $\pm$ 0.01 & 0.65 $\pm$ 0.01 \\ \cline{2-11}
 & FedProx                & 0.77 $\pm$ 0.01 & 0.68 $\pm$ 0.02 & 0.64 $\pm$ 0.01 & --- & --- & --- & 0.77 $\pm$ 0.01 & 0.68 $\pm$ 0.01 & 0.65 $\pm$ 0.01 \\ \cline{2-11}
 & FedAvgM                & 0.76 $\pm$ 0.01 & 0.68 $\pm$ 0.03 & 0.63 $\pm$ 0.01 & --- & --- & --- & 0.76 $\pm$ 0.01 & 0.67 $\pm$ 0.01 & 0.64 $\pm$ 0.01 \\ \cline{2-11}
 & FedAdagrad             & 0.77 $\pm$ 0.01 & 0.69 $\pm$ 0.01 & 0.69 $\pm$ 0.01 & --- & --- & --- & 0.77 $\pm$ 0.01 & 0.69 $\pm$ 0.01 & 0.69 $\pm$ 0.01 \\ \cline{2-11}
 & FedYogi                & 0.73 $\pm$ 0.01 & 0.70 $\pm$ 0.01 & 0.69 $\pm$ 0.01 & --- & --- & --- & 0.73 $\pm$ 0.01 & 0.70 $\pm$ 0.01 & 0.69 $\pm$ 0.01 \\ \cline{2-11}
 & FedAdam                & 0.71 $\pm$ 0.01 & 0.70 $\pm$ 0.01 & 0.70 $\pm$ 0.01 & --- & --- & --- & 0.71 $\pm$ 0.01 & 0.70 $\pm$ 0.01 & 0.69 $\pm$ 0.01 \\ \cline{2-11}
 & PoC                    & 0.77 $\pm$ 0.01 & 0.71 $\pm$ 0.03 & 0.65 $\pm$ 0.02 & --- & --- & --- & 0.77 $\pm$ 0.01 & 0.68 $\pm$ 0.01 & 0.65 $\pm$ 0.01 \\ \cline{2-11}
 & HACCS                  & 0.77 $\pm$ 0.01 & 0.72 $\pm$ 0.01 & 0.72 $\pm$ 0.01 & --- & --- & --- & 0.77 $\pm$ 0.01 & 0.74 $\pm$ 0.01 & 0.72 $\pm$ 0.01 \\ \cline{2-11}
 & FedCLS                 & 0.77 $\pm$ 0.01 & 0.71 $\pm$ 0.01 & 0.72 $\pm$ 0.01 & --- & --- & --- & 0.77 $\pm$ 0.01 & 0.73 $\pm$ 0.01 & 0.72 $\pm$ 0.02 \\ \cline{2-11}
 & CFL                    & 0.77 $\pm$ 0.01 & 0.76 $\pm$ 0.01 & 0.76 $\pm$ 0.01 & --- & --- & --- & 0.77 $\pm$ 0.01 & 0.76 $\pm$ 0.01 & 0.76 $\pm$ 0.01 \\ \cline{2-11}
 & FedSoft                & 0.77 $\pm$ 0.01 & 0.76 $\pm$ 0.01 & 0.75 $\pm$ 0.01 & --- & --- & --- & 0.77 $\pm$ 0.01 & 0.75 $\pm$ 0.01 & 0.74 $\pm$ 0.01 \\ \cline{2-11}
 & Clust-PSI-PFL (ours)   & \textbf{0.77 $\pm$ 0.01} & \textbf{0.89 $\pm$ 0.01} & \textbf{0.90 $\pm$ 0.01} & --- & --- & --- & \textbf{0.77 $\pm$ 0.01} & \textbf{0.97 $\pm$ 0.02} & \textbf{0.98 $\pm$ 0.01} \\ \hline

\multirow{13}{*}{Serengeti}
 & CL                     & \multicolumn{9}{c||}{0.96 $\pm$ 0.01} \\ \cline{2-11}
 & FedAvg                 & 0.78 $\pm$ 0.01 & 0.71 $\pm$ 0.01 & 0.69 $\pm$ 0.01 & 0.67 $\pm$ 0.02 & 0.56 $\pm$ 0.01 & 0.47 $\pm$ 0.01 & 0.78 $\pm$ 0.01 & 0.49 $\pm$ 0.01 & 0.13 $\pm$ 0.04 \\ \cline{2-11}
 & FedProx                & 0.78 $\pm$ 0.01 & 0.71 $\pm$ 0.01 & 0.69 $\pm$ 0.02 & 0.68 $\pm$ 0.01 & 0.59 $\pm$ 0.02 & 0.51 $\pm$ 0.01 & 0.77 $\pm$ 0.01 & 0.49 $\pm$ 0.01 & 0.18 $\pm$ 0.02 \\ \cline{2-11}
 & FedAvgM                & 0.81 $\pm$ 0.01 & 0.70 $\pm$ 0.02 & 0.64 $\pm$ 0.02 & 0.65 $\pm$ 0.02 & 0.52 $\pm$ 0.03 & 0.45 $\pm$ 0.03 & 0.81 $\pm$ 0.01 & 0.48 $\pm$ 0.01 & 0.15 $\pm$ 0.03 \\ \cline{2-11}
 & FedAdagrad             & 0.73 $\pm$ 0.01 & 0.68 $\pm$ 0.01 & 0.66 $\pm$ 0.01 & 0.65 $\pm$ 0.01 & 0.57 $\pm$ 0.01 & 0.50 $\pm$ 0.01 & 0.73 $\pm$ 0.01 & 0.50 $\pm$ 0.01 & 0.14 $\pm$ 0.02 \\ \cline{2-11}
 & FedYogi                & 0.77 $\pm$ 0.01 & 0.71 $\pm$ 0.01 & 0.71 $\pm$ 0.01 & 0.69 $\pm$ 0.01 & 0.60 $\pm$ 0.01 & 0.51 $\pm$ 0.02 & 0.77 $\pm$ 0.01 & 0.51 $\pm$ 0.01 & 0.15 $\pm$ 0.06 \\ \cline{2-11}
 & FedAdam                & 0.71 $\pm$ 0.01 & 0.65 $\pm$ 0.01 & 0.64 $\pm$ 0.01 & 0.62 $\pm$ 0.01 & 0.55 $\pm$ 0.01 & 0.48 $\pm$ 0.02 & 0.71 $\pm$ 0.01 & 0.46 $\pm$ 0.01 & 0.18 $\pm$ 0.03 \\ \cline{2-11}
 & PoC                    & 0.80 $\pm$ 0.01 & 0.68 $\pm$ 0.02 & 0.62 $\pm$ 0.03 & 0.50 $\pm$ 0.04 & 0.38 $\pm$ 0.12 & 0.37 $\pm$ 0.01 & 0.79 $\pm$ 0.01 & 0.28 $\pm$ 0.04 & 0.17 $\pm$ 0.01 \\ \cline{2-11}
 & HACCS                  & 0.78 $\pm$ 0.01 & 0.58 $\pm$ 0.01 & 0.59 $\pm$ 0.01 & 0.58 $\pm$ 0.04 & 0.44 $\pm$ 0.01 & 0.28 $\pm$ 0.08 & 0.79 $\pm$ 0.01 & 0.33 $\pm$ 0.03 & 0.22 $\pm$ 0.02 \\ \cline{2-11}
 & FedCLS                 & 0.79 $\pm$ 0.01 & 0.68 $\pm$ 0.02 & 0.59 $\pm$ 0.01 & 0.60 $\pm$ 0.01 & 0.34 $\pm$ 0.04 & 0.39 $\pm$ 0.03 & 0.78 $\pm$ 0.01 & 0.26 $\pm$ 0.01 & 0.18 $\pm$ 0.03 \\ \cline{2-11}
 & CFL                    & 0.82 $\pm$ 0.01 & 0.75 $\pm$ 0.01 & 0.70 $\pm$ 0.01 & 0.69 $\pm$ 0.01 & 0.60 $\pm$ 0.01 & 0.54 $\pm$ 0.01 & 0.82 $\pm$ 0.01 & 0.48 $\pm$ 0.01 & 0.37 $\pm$ 0.01 \\ \cline{2-11}
 & FedSoft                & 0.80 $\pm$ 0.01 & 0.72 $\pm$ 0.01 & 0.66 $\pm$ 0.01 & 0.59 $\pm$ 0.01 & 0.52 $\pm$ 0.02 & 0.36 $\pm$ 0.01 & 0.80 $\pm$ 0.01 & 0.31 $\pm$ 0.01 & 0.20 $\pm$ 0.01 \\ \cline{2-11}
 & Clust-PSI-PFL (ours)   & \textbf{0.82 $\pm$ 0.01} & \textbf{0.78 $\pm$ 0.02} & \textbf{0.72 $\pm$ 0.01} & \textbf{0.70 $\pm$ 0.01} & \textbf{0.76 $\pm$ 0.01} & \textbf{0.82 $\pm$ 0.01} & \textbf{0.82 $\pm$ 0.01} & \textbf{0.97 $\pm$ 0.01} & \textbf{0.98 $\pm$ 0.01} \\ \hline

\multirow{13}{*}{FMNIST}
 & CL                     & \multicolumn{9}{c||}{0.90 $\pm$ 0.02} \\ \cline{2-11}
 & FedAvg                 & 0.86 $\pm$ 0.01 & 0.83 $\pm$ 0.01 & \textbf{0.82 $\pm$ 0.01} & 0.81 $\pm$ 0.01 & 0.73 $\pm$ 0.01 & 0.68 $\pm$ 0.03 & 0.86 $\pm$ 0.01 & 0.74 $\pm$ 0.01 & 0.14 $\pm$ 0.04 \\ \cline{2-11}
 & FedProx                & 0.86 $\pm$ 0.01 & 0.83 $\pm$ 0.01 & 0.82 $\pm$ 0.01 & 0.80 $\pm$ 0.01 & 0.71 $\pm$ 0.02 & 0.64 $\pm$ 0.07 & 0.86 $\pm$ 0.01 & 0.73 $\pm$ 0.01 & 0.15 $\pm$ 0.07 \\ \cline{2-11}
 & FedAvgM                & 0.84 $\pm$ 0.01 & 0.82 $\pm$ 0.01 & 0.80 $\pm$ 0.01 & 0.77 $\pm$ 0.01 & 0.72 $\pm$ 0.01 & 0.57 $\pm$ 0.02 & 0.84 $\pm$ 0.01 & 0.74 $\pm$ 0.01 & 0.11 $\pm$ 0.01 \\ \cline{2-11}
 & FedAdagrad             & 0.84 $\pm$ 0.01 & 0.82 $\pm$ 0.01 & 0.80 $\pm$ 0.01 & 0.78 $\pm$ 0.01 & 0.72 $\pm$ 0.03 & 0.67 $\pm$ 0.03 & 0.84 $\pm$ 0.01 & 0.74 $\pm$ 0.01 & 0.12 $\pm$ 0.02 \\ \cline{2-11}
 & FedYogi                & 0.84 $\pm$ 0.01 & 0.81 $\pm$ 0.01 & 0.80 $\pm$ 0.01 & 0.77 $\pm$ 0.01 & 0.66 $\pm$ 0.03 & 0.66 $\pm$ 0.03 & 0.84 $\pm$ 0.01 & 0.74 $\pm$ 0.01 & 0.11 $\pm$ 0.02 \\ \cline{2-11}
 & FedAdam                & 0.82 $\pm$ 0.01 & 0.79 $\pm$ 0.01 & 0.77 $\pm$ 0.01 & 0.76 $\pm$ 0.01 & 0.68 $\pm$ 0.01 & 0.65 $\pm$ 0.02 & 0.82 $\pm$ 0.01 & 0.68 $\pm$ 0.01 & 0.14 $\pm$ 0.07 \\ \cline{2-11}
 & PoC                    & 0.77 $\pm$ 0.05 & 0.63 $\pm$ 0.02 & 0.44 $\pm$ 0.04 & 0.38 $\pm$ 0.03 & 0.26 $\pm$ 0.11 & 0.54 $\pm$ 0.05 & 0.14 $\pm$ 0.01 & 0.10 $\pm$ 0.03 & 0.10 $\pm$ 0.01 \\ \cline{2-11}
 & HACCS                  & 0.77 $\pm$ 0.05 & 0.79 $\pm$ 0.01 & 0.74 $\pm$ 0.01 & 0.73 $\pm$ 0.01 & 0.61 $\pm$ 0.02 & 0.61 $\pm$ 0.01 & 0.43 $\pm$ 0.14 & 0.16 $\pm$ 0.04 & 0.16 $\pm$ 0.04 \\ \cline{2-11}
 & FedCLS                 & 0.77 $\pm$ 0.01 & 0.79 $\pm$ 0.01 & 0.74 $\pm$ 0.01 & 0.72 $\pm$ 0.01 & 0.58 $\pm$ 0.05 & 0.53 $\pm$ 0.03 & 0.42 $\pm$ 0.14 & 0.10 $\pm$ 0.02 & 0.10 $\pm$ 0.03 \\ \cline{2-11}
 & CFL                    & 0.81 $\pm$ 0.01 & 0.79 $\pm$ 0.01 & 0.74 $\pm$ 0.01 & 0.77 $\pm$ 0.01 & 0.62 $\pm$ 0.01 & 0.64 $\pm$ 0.01 & 0.81 $\pm$ 0.01 & 0.61 $\pm$ 0.02 & 0.59 $\pm$ 0.02 \\ \cline{2-11}
 & FedSoft                & 0.81 $\pm$ 0.01 & 0.74 $\pm$ 0.02 & 0.71 $\pm$ 0.01 & 0.71 $\pm$ 0.01 & 0.46 $\pm$ 0.01 & 0.46 $\pm$ 0.04 & 0.81 $\pm$ 0.01 & 0.33 $\pm$ 0.02 & 0.28 $\pm$ 0.02 \\ \cline{2-11}
 & Clust-PSI-PFL (ours)   & \textbf{0.86 $\pm$ 0.01} & \textbf{0.83 $\pm$ 0.01} & 0.80 $\pm$ 0.01 & \textbf{0.83 $\pm$ 0.01} & \textbf{0.84 $\pm$ 0.01} & \textbf{0.83 $\pm$ 0.01} & \textbf{0.86 $\pm$ 0.01} & \textbf{0.98 $\pm$ 0.01} & \textbf{0.98 $\pm$ 0.01} \\ \hline

\multirow{13}{*}{CIFAR10}
 & CL                     & \multicolumn{9}{c||}{0.84 $\pm$ 0.02} \\ \cline{2-11}
 & FedAvg                 & 0.74 $\pm$ 0.01 & 0.68 $\pm$ 0.03 & 0.52 $\pm$ 0.05 & 0.45 $\pm$ 0.08 & 0.18 $\pm$ 0.02 & 0.16 $\pm$ 0.03 & 0.75 $\pm$ 0.01 & 0.13 $\pm$ 0.01 & 0.10 $\pm$ 0.01 \\ \cline{2-11}
 & FedProx                & 0.74 $\pm$ 0.01 & 0.66 $\pm$ 0.03 & 0.59 $\pm$ 0.02 & 0.41 $\pm$ 0.02 & 0.13 $\pm$ 0.02 & 0.15 $\pm$ 0.03 & 0.75 $\pm$ 0.01 & 0.14 $\pm$ 0.02 & 0.10 $\pm$ 0.01 \\ \cline{2-11}
 & FedAvgM                & 0.75 $\pm$ 0.01 & \textbf{0.70 $\pm$ 0.02} & 0.53 $\pm$ 0.05 & 0.42 $\pm$ 0.05 & 0.16 $\pm$ 0.01 & 0.12 $\pm$ 0.03 & 0.75 $\pm$ 0.01 & 0.10 $\pm$ 0.01 & 0.10 $\pm$ 0.01 \\ \cline{2-11}
 & FedAdagrad             & 0.63 $\pm$ 0.02 & 0.44 $\pm$ 0.06 & 0.40 $\pm$ 0.07 & 0.25 $\pm$ 0.01 & 0.17 $\pm$ 0.07 & 0.11 $\pm$ 0.01 & 0.61 $\pm$ 0.01 & 0.13 $\pm$ 0.04 & 0.08 $\pm$ 0.03 \\ \cline{2-11}
 & FedYogi                & 0.44 $\pm$ 0.05 & 0.29 $\pm$ 0.11 & 0.27 $\pm$ 0.04 & 0.33 $\pm$ 0.08 & 0.10 $\pm$ 0.01 & 0.14 $\pm$ 0.01 & 0.51 $\pm$ 0.06 & 0.13 $\pm$ 0.02 & 0.10 $\pm$ 0.02 \\ \cline{2-11}
 & FedAdam                & 0.53 $\pm$ 0.07 & 0.38 $\pm$ 0.03 & 0.17 $\pm$ 0.06 & 0.13 $\pm$ 0.05 & 0.10 $\pm$ 0.01 & 0.10 $\pm$ 0.03 & 0.49 $\pm$ 0.08 & 0.09 $\pm$ 0.01 & 0.10 $\pm$ 0.03 \\ \cline{2-11}
 & PoC                    & 0.60 $\pm$ 0.02 & 0.49 $\pm$ 0.01 & 0.44 $\pm$ 0.02 & 0.50 $\pm$ 0.02 & 0.33 $\pm$ 0.01 & 0.38 $\pm$ 0.01 & 0.55 $\pm$ 0.02 & 0.23 $\pm$ 0.03 & 0.12 $\pm$ 0.01 \\ \cline{2-11}
 & HACCS                  & 0.76 $\pm$ 0.01 & 0.66 $\pm$ 0.03 & 0.62 $\pm$ 0.02 & 0.63 $\pm$ 0.03 & 0.49 $\pm$ 0.01 & 0.46 $\pm$ 0.02 & 0.73 $\pm$ 0.01 & 0.31 $\pm$ 0.05 & 0.11 $\pm$ 0.02 \\ \cline{2-11}
 & FedCLS                 & 0.76 $\pm$ 0.01 & 0.65 $\pm$ 0.03 & 0.60 $\pm$ 0.03 & 0.63 $\pm$ 0.01 & 0.47 $\pm$ 0.01 & 0.46 $\pm$ 0.01 & 0.72 $\pm$ 0.02 & 0.29 $\pm$ 0.04 & 0.14 $\pm$ 0.03 \\ \cline{2-11}
 & CFL                    & 0.77 $\pm$ 0.01 & 0.66 $\pm$ 0.03 & 0.59 $\pm$ 0.01 & 0.64 $\pm$ 0.02 & 0.45 $\pm$ 0.02 & 0.45 $\pm$ 0.01 & 0.73 $\pm$ 0.01 & 0.35 $\pm$ 0.01 & 0.19 $\pm$ 0.02 \\ \cline{2-11}
 & FedSoft                & 0.76 $\pm$ 0.01 & 0.59 $\pm$ 0.03 & 0.53 $\pm$ 0.03 & 0.60 $\pm$ 0.02 & 0.32 $\pm$ 0.01 & 0.34 $\pm$ 0.03 & 0.71 $\pm$ 0.01 & 0.23 $\pm$ 0.03 & 0.13 $\pm$ 0.01 \\ \cline{2-11}
 & Clust-PSI-PFL (ours)   & \textbf{0.77 $\pm$ 0.01} & 0.68 $\pm$ 0.02 & \textbf{0.63 $\pm$ 0.01} & \textbf{0.64 $\pm$ 0.01} & \textbf{0.71 $\pm$ 0.02} & \textbf{0.85 $\pm$ 0.01} & \textbf{0.75 $\pm$ 0.01} & \textbf{0.94 $\pm$ 0.01} & \textbf{0.98 $\pm$ 0.01} \\ \hline

\multirow{13}{*}{Sent140}
 & CL                     & \multicolumn{9}{c||}{0.74 $\pm$ 0.01} \\ \cline{2-11}
 & FedAvg                 & 0.68 $\pm$ 0.01 & 0.60 $\pm$ 0.01 & 0.68 $\pm$ 0.01 & --- & --- & --- & 0.71 $\pm$ 0.01 & 0.53 $\pm$ 0.03 & 0.56 $\pm$ 0.05 \\ \cline{2-11}
 & FedProx                & 0.68 $\pm$ 0.02 & 0.58 $\pm$ 0.03 & 0.67 $\pm$ 0.02 & --- & --- & --- & 0.71 $\pm$ 0.01 & 0.60 $\pm$ 0.02 & 0.55 $\pm$ 0.05 \\ \cline{2-11}
 & FedAvgM                & 0.70 $\pm$ 0.01 & 0.64 $\pm$ 0.07 & 0.50 $\pm$ 0.01 & --- & --- & --- & 0.72 $\pm$ 0.01 & 0.51 $\pm$ 0.01 & 0.53 $\pm$ 0.05 \\ \cline{2-11}
 & FedAdagrad             & 0.68 $\pm$ 0.01 & 0.66 $\pm$ 0.01 & 0.62 $\pm$ 0.01 & --- & --- & --- & 0.69 $\pm$ 0.01 & 0.58 $\pm$ 0.02 & 0.56 $\pm$ 0.03 \\ \cline{2-11}
 & FedYogi                & 0.59 $\pm$ 0.01 & 0.66 $\pm$ 0.03 & 0.57 $\pm$ 0.03 & --- & --- & --- & 0.71 $\pm$ 0.01 & 0.59 $\pm$ 0.03 & 0.55 $\pm$ 0.03 \\ \cline{2-11}
 & FedAdam                & 0.70 $\pm$ 0.01 & 0.66 $\pm$ 0.01 & 0.66 $\pm$ 0.01 & --- & --- & --- & 0.70 $\pm$ 0.01 & 0.60 $\pm$ 0.02 & 0.58 $\pm$ 0.04 \\ \cline{2-11}
 & PoC                    & 0.68 $\pm$ 0.01 & 0.52 $\pm$ 0.02 & 0.50 $\pm$ 0.01 & --- & --- & --- & 0.54 $\pm$ 0.02 & 0.50 $\pm$ 0.01 & 0.50 $\pm$ 0.01 \\ \cline{2-11}
 & HACCS                  & 0.68 $\pm$ 0.01 & 0.64 $\pm$ 0.02 & 0.53 $\pm$ 0.02 & --- & --- & --- & 0.72 $\pm$ 0.01 & 0.51 $\pm$ 0.01 & 0.51 $\pm$ 0.02 \\ \cline{2-11}
 & FedCLS                 & 0.67 $\pm$ 0.01 & 0.62 $\pm$ 0.01 & 0.53 $\pm$ 0.01 & --- & --- & --- & 0.72 $\pm$ 0.01 & 0.51 $\pm$ 0.01 & 0.49 $\pm$ 0.01 \\ \cline{2-11}
 & CFL                    & 0.71 $\pm$ 0.01 & 0.61 $\pm$ 0.01 & 0.56 $\pm$ 0.01 & --- & --- & --- & 0.72 $\pm$ 0.01 & 0.53 $\pm$ 0.01 & 0.53 $\pm$ 0.01 \\ \cline{2-11}
 & FedSoft                & 0.71 $\pm$ 0.01 & 0.59 $\pm$ 0.02 & 0.54 $\pm$ 0.02 & --- & --- & --- & 0.71 $\pm$ 0.01 & 0.52 $\pm$ 0.02 & 0.51 $\pm$ 0.02 \\ \cline{2-11}
 & Clust-PSI-PFL (ours)   & \textbf{0.71 $\pm$ 0.01} & \textbf{0.81 $\pm$ 0.02} & \textbf{0.84 $\pm$ 0.01} & --- & --- & --- & \textbf{0.72 $\pm$ 0.01} & \textbf{0.97 $\pm$ 0.02} & \textbf{0.98 $\pm$ 0.01} \\ \hline

\multirow{13}{*}{\makecell{Amazon \\ reviews}}
 & CL                     & \multicolumn{9}{c||}{0.77 $\pm$ 0.02} \\ \cline{2-11}
 & FedAvg                 & 0.73 $\pm$ 0.01 & 0.51 $\pm$ 0.01 & 0.34 $\pm$ 0.02 & 0.42 $\pm$ 0.05 & 0.38 $\pm$ 0.05 & --- & 0.73 $\pm$ 0.01 & 0.24 $\pm$ 0.05 & 0.30 $\pm$ 0.03 \\ \cline{2-11}
 & FedProx                & 0.73 $\pm$ 0.01 & 0.61 $\pm$ 0.01 & 0.52 $\pm$ 0.01 & 0.49 $\pm$ 0.01 & 0.37 $\pm$ 0.02 & --- & 0.73 $\pm$ 0.01 & 0.28 $\pm$ 0.02 & 0.25 $\pm$ 0.01 \\ \cline{2-11}
 & FedAvgM                & 0.69 $\pm$ 0.01 & 0.66 $\pm$ 0.01 & 0.51 $\pm$ 0.03 & 0.31 $\pm$ 0.04 & 0.26 $\pm$ 0.01 & --- & 0.68 $\pm$ 0.01 & 0.36 $\pm$ 0.01 & 0.27 $\pm$ 0.03 \\ \cline{2-11}
 & FedAdagrad             & 0.54 $\pm$ 0.01 & 0.48 $\pm$ 0.01 & 0.34 $\pm$ 0.02 & 0.26 $\pm$ 0.01 & 0.26 $\pm$ 0.01 & --- & 0.53 $\pm$ 0.01 & 0.27 $\pm$ 0.02 & 0.27 $\pm$ 0.02 \\ \cline{2-11}
 & FedYogi                & 0.71 $\pm$ 0.01 & 0.64 $\pm$ 0.01 & 0.59 $\pm$ 0.01 & 0.45 $\pm$ 0.02 & 0.39 $\pm$ 0.01 & --- & 0.71 $\pm$ 0.01 & 0.39 $\pm$ 0.04 & 0.43 $\pm$ 0.05 \\ \cline{2-11}
 & FedAdam                & 0.66 $\pm$ 0.01 & 0.55 $\pm$ 0.01 & 0.38 $\pm$ 0.01 & 0.40 $\pm$ 0.07 & 0.38 $\pm$ 0.06 & --- & 0.66 $\pm$ 0.01 & 0.34 $\pm$ 0.06 & 0.32 $\pm$ 0.06 \\ \cline{2-11}
 & PoC                    & 0.66 $\pm$ 0.02 & 0.44 $\pm$ 0.01 & 0.34 $\pm$ 0.01 & 0.49 $\pm$ 0.02 & 0.36 $\pm$ 0.03 & --- & 0.66 $\pm$ 0.02 & 0.22 $\pm$ 0.04 & 0.20 $\pm$ 0.05 \\ \cline{2-11}
 & HACCS                  & 0.70 $\pm$ 0.01 & 0.69 $\pm$ 0.01 & 0.66 $\pm$ 0.01 & 0.63 $\pm$ 0.01 & 0.57 $\pm$ 0.01 & --- & 0.69 $\pm$ 0.01 & 0.46 $\pm$ 0.02 & 0.28 $\pm$ 0.03 \\ \cline{2-11}
 & FedCLS                 & 0.69 $\pm$ 0.01 & 0.69 $\pm$ 0.01 & 0.66 $\pm$ 0.01 & 0.65 $\pm$ 0.01 & 0.56 $\pm$ 0.01 & --- & 0.69 $\pm$ 0.01 & 0.49 $\pm$ 0.06 & 0.25 $\pm$ 0.01 \\ \cline{2-11}
 & CFL                    & 0.73 $\pm$ 0.01 & \textbf{0.70 $\pm$ 0.01} & 0.59 $\pm$ 0.01 & 0.67 $\pm$ 0.01 & 0.70 $\pm$ 0.01 & --- & 0.74 $\pm$ 0.01 & 0.73 $\pm$ 0.01 & 0.73 $\pm$ 0.01 \\ \cline{2-11}
 & FedSoft                & 0.32 $\pm$ 0.02 & 0.28 $\pm$ 0.01 & 0.26 $\pm$ 0.01 & 0.25 $\pm$ 0.01 & 0.25 $\pm$ 0.02 & --- & 0.32 $\pm$ 0.01 & 0.25 $\pm$ 0.01 & 0.24 $\pm$ 0.02 \\ \cline{2-11}
 & Clust-PSI-PFL (ours)   & \textbf{0.73 $\pm$ 0.01} & 0.68 $\pm$ 0.02 & \textbf{0.66 $\pm$ 0.01} & \textbf{0.70 $\pm$ 0.01} & \textbf{0.79 $\pm$ 0.01} & --- & \textbf{0.74 $\pm$ 0.01} & \textbf{0.85 $\pm$ 0.01} & \textbf{0.98 $\pm$ 0.01} \\ \hline

\end{tabular}}
\label{tab:global_dirichlet_similarity_by_dataset}
\end{table*}

To summarize results across datasets, Table~\ref{tab:global_dirichlet_similarity_by_dataset} reports mean test accuracy ($\pm$ standard deviation over five seeds) for Clust-PSI-PFL and all baselines under both partition protocols (Dirichlet and Similarity) and across modalities. This table enables a direct, cross-dataset comparison of behavior under varying levels of non-IID data. For brevity, we show results for $K=100$ clients; qualitatively similar trends hold for $K\in \{10,50\}$.

Across all six datasets and both partition protocols, Clust-PSI-PFL (ours) attains the best accuracy in the vast majority of settings. Its advantage is most pronounced in the more pathological non-IID regimes (Dirichlet with small $\alpha$ and Similarity with small $S$) where it delivers large gains over strong baselines (FedAvg, FedAvgM, HACCS, CFL). In near-IID conditions ($\alpha=50$, $S=1$), it matches the top performers, indicating no loss from clustering when non-IID data is low. Reported standard deviations are small, evidencing stable training. Overall, this analysis confirms that PSI-driven clustering yields consistently superior global accuracy across modalities and non-IID data levels.

\vspace{0.5em}
\noindent \vspace{0.1em}\mybox{gray}{\observation{
Clust-PSI-PFL consistently attains higher global accuracy than competing baselines across the full spectrum of non-IID data scenarios.
}}
\vspace{0.5em}

\paragraph{\textbf{Local-level Test Clust-PSI-PFL Behavior}}

Local-level analysis examines the performance of each client’s model before and after federated aggregation. This perspective surfaces variability arising from non-IID data distributions, computational limitations, and personalization requirements. By evaluating outcomes at the client granularity, we can quantify the impact of FL on individual nodes, assess fairness, and detect disparities in model quality across clients.

EQ3 is addressed using the evidence in Fig.~\ref{fig:local_acc_both}, which plots the empirical cumulative distribution function (ECDF) of local accuracy for Clust-PSI-PFL and all baselines across varying $\alpha$ and $S$. The ECDF of our method lies consistently closer to the perfect-accuracy target, indicating improved client fairness. At $\alpha=0.3$, Clust-PSI-PFL attains $AD=0.17$, versus $AD \sim 0.27$ for CFL (37\% of relative reduction) and $AD \sim 0.43$ for FedAvgM. Under the pathological non-IID scenario ($S=0$), our approach achieves $AD=0.01$ from the unit-accuracy target with $SDAD=0.02$, whereas CFL yields $AD=0.27$ and $SDAD=0.28$. These results show that Clust-PSI-PFL concentrates client performance nearer to the ideal with markedly lower dispersion.

\begin{table*}[t]
\centering
\caption{Local average distance (AD $\pm$ SDAD) for baselines under both partition protocols on all the datasets for $K=100$. Each simulation was run for 5 seeds (trials). Showing only the best baselines from Table~\ref{tab:global_dirichlet_similarity_by_dataset}. ``---'' denotes infeasible partitions. Best performing shown in bold.}
\resizebox{\textwidth}{!}{
\begin{tabular}{||c|l|cccccc|ccc||}
\hline
& & \multicolumn{6}{c|}{\textbf{Dirichlet}} & \multicolumn{3}{c||}{\textbf{Similarity}} \\ \cline{3-11}
\textbf{Dataset} & \textbf{Baseline} & $\alpha=50$ & $\alpha=0.7$ & $\alpha=0.3$ & $\alpha=0.2$ & $\alpha=0.09$ & $\alpha=0.05$ & $S=1$ & $S=0.03$ & $S=0$ \\ \hline\hline

\multirow{5}{*}{ACSIncome}
 & FedAvg               & 0.22 $\pm$ 0.01 & 0.36 $\pm$ 0.30 & 0.43 $\pm$ 0.39 & --- & --- & --- & 0.22 $\pm$ 0.01 & 0.31 $\pm$ 0.35 & 0.35 $\pm$ 0.44 \\ \cline{2-11}
 & FedAvgM              & 0.23 $\pm$ 0.01 & 0.35 $\pm$ 0.26 & 0.43 $\pm$ 0.40 & --- & --- & --- & 0.23 $\pm$ 0.01 & 0.31 $\pm$ 0.32 & 0.35 $\pm$ 0.39 \\ \cline{2-11}
 & HACCS                & 0.22 $\pm$ 0.0 & 0.30 $\pm$ 0.22 & 0.32 $\pm$ 0.25 & --- & --- & --- & 0.22 $\pm$ 0.01 & 0.27 $\pm$ 0.27 & 0.29 $\pm$ 0.34 \\ \cline{2-11}
 & CFL                  & 0.22 $\pm$ 0.01 & 0.28 $\pm$ 0.18 & 0.27 $\pm$ 0.17 & --- & --- & --- & 0.22 $\pm$ 0.01 & 0.26 $\pm$ 0.24 & 0.27 $\pm$ 0.28 \\ \cline{2-11}
 & Clust-PSI-PFL (ours) & \textbf{0.22 $\pm$ 0.01} & \textbf{0.14 $\pm$ 0.10} & \textbf{0.17 $\pm$ 0.18} & \textbf{---} & \textbf{---} & \textbf{---} & \textbf{0.22 $\pm$ 0.01} & \textbf{0.02 $\pm$ 0.07} & \textbf{0.01 $\pm$ 0.02} \\ \hline

\multirow{5}{*}{Serengeti}
 & FedAvg               & 0.18 $\pm$ 0.05 & 0.28 $\pm$ 0.11 & \textbf{0.27 $\pm$ 0.12} & \textbf{0.30 $\pm$ 0.13} & 0.43 $\pm$ 0.18 & 0.54 $\pm$ 0.19 & 0.18 $\pm$ 0.05 & 0.46 $\pm$ 0.16 & 0.82 $\pm$ 0.18 \\ \cline{2-11}
 & FedAvgM              & 0.15 $\pm$ 0.03 & 0.28 $\pm$ 0.12 & 0.36 $\pm$ 0.13 & 0.35 $\pm$ 0.10 & 0.48 $\pm$ 0.21 & 0.60 $\pm$ 0.16 & 0.16 $\pm$ 0.03 & 0.50 $\pm$ 0.15 & 0.84 $\pm$ 0.30 \\ \cline{2-11}
 & HACCS                & 0.17 $\pm$ 0.05 & 0.40 $\pm$ 0.12 & 0.40 $\pm$ 0.12 & 0.40 $\pm$ 0.10 & 0.55 $\pm$ 0.14 & 0.71 $\pm$ 0.14 & 0.17 $\pm$ 0.04 & 0.60 $\pm$ 0.18 & 0.78 $\pm$ 0.16 \\ \cline{2-11}
 & CFL                  & 0.14 $\pm$ 0.05 & 0.25 $\pm$ 0.07 & 0.29 $\pm$ 0.08 & 0.27 $\pm$ 0.10 & 0.41 $\pm$ 0.10 & 0.44 $\pm$ 0.15 & 0.14 $\pm$ 0.05 & 0.47 $\pm$ 0.18 & 0.63 $\pm$ 0.21 \\ \cline{2-11}
 & Clust-PSI-PFL (ours) & \textbf{0.14 $\pm$ 0.04} & \textbf{0.21 $\pm$ 0.09} & 0.31 $\pm$ 0.16 & 0.30 $\pm$ 0.20 & \textbf{0.30 $\pm$ 0.31} & \textbf{0.28 $\pm$ 0.32} & \textbf{0.19 $\pm$ 0.05} & \textbf{0.03 $\pm$ 0.04} & \textbf{0.01 $\pm$ 0.08} \\ \hline

\multirow{5}{*}{FMNIST}
 & FedAvg               & 0.12 $\pm$ 0.02 & \textbf{0.14 $\pm$ 0.06} & \textbf{0.16 $\pm$ 0.10} & 0.18 $\pm$ 0.12 & 0.28 $\pm$ 0.24 & 0.35 $\pm$ 0.23 & 0.12 $\pm$ 0.02 & 0.24 $\pm$ 0.20 & 0.84 $\pm$ 0.19 \\ \cline{2-11}
 & FedAvgM              & 0.14 $\pm$ 0.02 & 0.17 $\pm$ 0.06 & 0.19 $\pm$ 0.10 & 0.23 $\pm$ 0.14 & 0.29 $\pm$ 0.21 & 0.45 $\pm$ 0.24 & 0.14 $\pm$ 0.02 & 0.25 $\pm$ 0.21 & 0.89 $\pm$ 0.20 \\ \cline{2-11}
 & HACCS                & 0.19 $\pm$ 0.04 & 0.18 $\pm$ 0.07 & 0.21 $\pm$ 0.11 & 0.24 $\pm$ 0.15 & 0.41 $\pm$ 0.24 & 0.45 $\pm$ 0.25 & 0.46 $\pm$ 0.14 & 0.85 $\pm$ 0.22 & 0.84 $\pm$ 0.23 \\ \cline{2-11}
 & CFL                  & 0.17 $\pm$ 0.03 & 0.21 $\pm$ 0.07 & 0.24 $\pm$ 0.12 & 0.22 $\pm$ 0.15 & 0.43 $\pm$ 0.19 & 0.40 $\pm$ 0.23 & 0.17 $\pm$ 0.03 & 0.44 $\pm$ 0.30 & 0.51 $\pm$ 0.24 \\ \cline{2-11}
 & Clust-PSI-PFL (ours) & \textbf{0.12 $\pm$ 0.03} & 0.16 $\pm$ 0.11 & 0.18 $\pm$ 0.15 & \textbf{0.16 $\pm$ 0.19} & \textbf{0.23 $\pm$ 0.34} & \textbf{0.20 $\pm$ 0.30} & \textbf{0.12 $\pm$ 0.02} & \textbf{0.01 $\pm$ 0.03} & \textbf{0.01 $\pm$ 0.01} \\ \hline

\multirow{5}{*}{CIFAR10}
 & FedAvg               & 0.23 $\pm$ 0.03 & 0.31 $\pm$ 0.06 & 0.48 $\pm$ 0.07 & 0.57 $\pm$ 0.16 & 0.84 $\pm$ 0.17 & 0.80 $\pm$ 0.24 & 0.23 $\pm$ 0.03 & 0.82 $\pm$ 0.18 & 0.89 $\pm$ 0.13 \\ \cline{2-11}
 & FedAvgM              & 0.23 $\pm$ 0.03 & \textbf{0.2 $\pm$ 0.06} & 0.45 $\pm$ 0.10 & 0.60 $\pm$ 0.17 & 0.83 $\pm$ 0.13 & 0.82 $\pm$ 0.18 & \textbf{0.22 $\pm$ 0.03} & 0.85 $\pm$ 0.15 & 0.90 $\pm$ 0.15 \\ \cline{2-11}
 & HACCS                & 0.22 $\pm$ 0.03 & 0.33 $\pm$ 0.05 & 0.38 $\pm$ 0.08 & 0.37 $\pm$ 0.12 & 0.51 $\pm$ 0.13 & 0.52 $\pm$ 0.18 & 0.24 $\pm$ 0.04 & 0.67 $\pm$ 0.18 & 0.88 $\pm$ 0.14 \\ \cline{2-11}
 & CFL                  & 0.21 $\pm$ 0.03 & 0.34 $\pm$ 0.05 & 0.43 $\pm$ 0.09 & 0.36 $\pm$ 0.26 & 0.56 $\pm$ 0.11 & 0.56 $\pm$ 0.16 & 0.24 $\pm$ 0.04 & 0.68 $\pm$ 0.18 & 0.85 $\pm$ 0.17 \\ \cline{2-11}
 & Clust-PSI-PFL (ours) & \textbf{0.21 $\pm$ 0.05} & 0.34 $\pm$ 0.11 & \textbf{0.37 $\pm$ 0.23} & \textbf{0.36 $\pm$ 0.10} & \textbf{0.45 $\pm$ 0.38} & \textbf{0.33 $\pm$ 0.35} & 0.26 $\pm$ 0.04 & \textbf{0.04 $\pm$ 0.13} & \textbf{0.01 $\pm$ 0.06} \\ \hline

\multirow{5}{*}{Sent140}
 & FedAvg               & 0.29 $\pm$ 0.02 & 0.39 $\pm$ 0.22 & 0.34 $\pm$ 0.15 & --- & --- & --- & 0.27 $\pm$ 0.01 & 0.43 $\pm$ 0.28 & 0.44 $\pm$ 0.27 \\ \cline{2-11}
 & FedAvgM              & 0.28 $\pm$ 0.02 & 0.36 $\pm$ 0.11 & 0.48 $\pm$ 0.12 & --- & --- & --- & 0.27 $\pm$ 0.01 & 0.48 $\pm$ 0.16 & 0.47 $\pm$ 0.04 \\ \cline{2-11}
 & HACCS                & 0.29 $\pm$ 0.02 & 0.34 $\pm$ 0.03 & 0.44 $\pm$ 0.22 & --- & --- & --- & 0.27 $\pm$ 0.01 & 0.49 $\pm$ 0.25 & 0.49 $\pm$ 0.10 \\ \cline{2-11}
 & CFL                  & 0.28 $\pm$ 0.01 & 0.38 $\pm$ 0.08 & 0.44 $\pm$ 0.24 & --- & --- & --- & 0.27 $\pm$ 0.01 & 0.49 $\pm$ 0.39 & 0.49 $\pm$ 0.41 \\ \cline{2-11}
 & Clust-PSI-PFL (ours) & \textbf{0.27 $\pm$ 0.01} & \textbf{0.22 $\pm$ 0.13} & \textbf{0.22 $\pm$ 0.24} & \textbf{---} & \textbf{---} & \textbf{---} & \textbf{0.27 $\pm$ 0.01} & \textbf{0.02 $\pm$ 0.01} & \textbf{0.01 $\pm$ 0.01} \\ \hline

\multirow{5}{*}{\makecell{Amazon \\ reviews}}
 & FedAvg               & 0.26 $\pm$ 0.04 & 0.49 $\pm$ 0.12 & 0.67 $\pm$ 0.14 & 0.59 $\pm$ 0.20 & 0.67 $\pm$ 0.16 & --- & 0.26 $\pm$ 0.04 & 0.71 $\pm$ 0.21 & 0.70 $\pm$ 0.23 \\ \cline{2-11}
 & FedAvgM              & 0.32 $\pm$ 0.05 & 0.34 $\pm$ 0.06 & 0.49 $\pm$ 0.07 & 0.67 $\pm$ 0.18 & 0.77 $\pm$ 0.20 & --- & 0.33 $\pm$ 0.05 & 0.64 $\pm$ 0.18 & 0.72 $\pm$ 0.32 \\ \cline{2-11}
 & HACCS                & 0.29 $\pm$ 0.03 & 0.30 $\pm$ 0.05 & 0.34 $\pm$ 0.04 & 0.37 $\pm$ 0.09 & 0.46 $\pm$ 0.10 & --- & 0.30 $\pm$ 0.03 & 0.54 $\pm$ 0.29 & 0.71 $\pm$ 0.33 \\ \cline{2-11}
 & CFL                  & 0.26 $\pm$ 0.03 & \textbf{0.30 $\pm$ 0.05} & 0.31 $\pm$ 0.07 & 0.36 $\pm$ 0.17 & 0.39 $\pm$ 0.17 & --- & 0.26 $\pm$ 0.03 & 0.37 $\pm$ 0.20 & 0.40 $\pm$ 0.21 \\ \cline{2-11}
 & Clust-PSI-PFL (ours) & \textbf{0.26 $\pm$ 0.07} & 0.32 $\pm$ 0.18 & \textbf{0.31 $\pm$ 0.22} & \textbf{0.36 $\pm$ 0.27} & \textbf{0.34 $\pm$ 0.31} & \textbf{---} & \textbf{0.26 $\pm$ 0.06} & \textbf{0.14 $\pm$ 0.23} & \textbf{0.03 $\pm$ 0.10} \\ \hline

\end{tabular}}
\label{tab:local_dirichlet_similarity_by_dataset}
\end{table*}

To summarize local‐accuracy results across datasets, Table~\ref{tab:local_dirichlet_similarity_by_dataset} reports the average distance from the perfect model (AD) with its dispersion (SDAD) as $AD \pm SDAD,$ for Clust-PSI-PFL and a focused set of strong baselines. Specifically, we include one representative per PFL family—regularization (FedAvgM), selection (HACCS), and clustering (CFL)—together with the FedAvg reference, selected because these methods achieved the highest global accuracies in Table~\ref{tab:global_dirichlet_similarity_by_dataset}. The columns mirror the non-IID settings for both Dirichlet and Similarity protocols, enabling cross‐dataset and cross‐modality comparisons as non-IID data varies (lower $AD/SDAD$ is better). For brevity, results are shown for $K=100$ clients; qualitatively similar trends were observed for $K\in \{10,50\}$.

Table~\ref{tab:local_dirichlet_similarity_by_dataset} shows that Clust-PSI-PFL attains the smallest \(AD\), and typically the smallest \(SDAD\), across nearly all datasets, with gains widening as non-IID intensifies. Under severe Similarity skew \(\left(S\in\{0,0.03\}\right)\) it reaches near-zero \(AD\) while baselines remain far (e.g., ACSIncome at \(S=0\): \(0.01\!\pm\!0.02\) vs.\ \(0.27\)–\(0.35\); FMNIST: \(0.01\!\pm\!0.01\) vs.\ \(0.51\)–\(0.89\)). For Dirichlet at moderate non-IID \(\left(\alpha\in\{0.7,0.3\}\right)\), it also leads (e.g., ACSIncome \(\alpha=0.7\): \(0.14\!\pm\!0.10\) vs.\ \(0.28\)–\(0.36\)); at \(\alpha\!\sim\!0.2\) CFL can be slightly ahead on some vision sets, but our method dominates as non-IID strengthens and in Similarity. Near-IID (\(\alpha=50\), \(S=1\)), all methods are comparable, indicating no clustering penalty. Overall, consistently smaller \(AD\)/\(SDAD\) for Clust-PSI-PFL reflect stronger client fairness and stability.

\vspace{0.5em}
\noindent \vspace{0.1em} \mybox{gray}{\observation{Clust-PSI-PFL promotes client fairness, attaining an improvement of 37\% compared to the baselines and under highly non-IID conditions.}}

\section{Limitations of Clust-PSI-PFL}
\label{sec:limitations}
While Clust-PSI-PFL shows consistent gains across datasets and non-IID scenarios, our current implementation does not incorporate privacy-preserving mechanisms such as DP or MPC. Integrating MPC-based secure aggregation does not change model performance (it adds only runtime/communication overhead), whereas DP may affect performance, with the impact governed by the budget $\epsilon$. Providing formal privacy guarantees is left to DP/MPC-enhanced variants. 


Second, the present version of Clust-PSI-PFL focuses on label skew: PSI features are derived from class frequency, and clients are clustered by similarity in label proportions. Other non-IID data types (e.g., attribute skew, quantity imbalance) are not explicitly modeled; thus, when labels are balanced, but attribute distributions or data volumes differ, clusters may be suboptimal. We view this as a scoped design choice rather than a fundamental limitation: the PSI machinery can be extended to discretized attributes, marginals, or joint label–attribute bins, and incorporate sample counts. 

Finally, when some clients have extremely small local datasets, their label histograms (and thus PSI) can be noisy. A simple mitigation is to down-weight such clients during clustering or append $n_i$ as an auxiliary clustering feature. A systematic evaluation of such extremely small-client regimes is an important direction for future work.

\section{Conclusion and Future Work}
\label{sec:conclusion}
In this study, we present Clust-PSI-PFL, a PFL framework that leverages the PSI as a principled metric for client clustering under non-IID data conditions. We demonstrate that PSI reliably quantifies distributional non-IID data across tabular, image, and text modalities, enabling the formation of distributionally coherent client groups and, in turn, yielding substantial gains in both global and local performance, even in highly non-IID regimes. Relative to state-of-the-art baselines, Clust-PSI-PFL achieves up to 18\% higher global accuracy while enhancing client fairness, resulting in a relative improvement of 37\%. These results position Clust-PSI-PFL as a practical mechanism for robust decentralized learning in non-IID data federated environments.

In future work, we will examine the applicability of PSI beyond label skew to other forms of data skew, such as attribute skew and quantity skew. Moreover, integrating formal privacy mechanisms, such as DP\cite{erlingsson2014rappor} and MPC\cite{bohler2021secure}, to strengthen the confidentiality of our FL pipeline lies outside the scope of this study and remains a promising direction for follow-up research.

\bibliographystyle{IEEEtran}
\bibliography{references}

\twocolumn[%
{\begin{center}
\Huge
Appendix: Artifact Description/Artifact Evaluation
\end{center}}
]


\appendixAD

\section{Overview of Contributions and Artifacts}

\subsection{Paper's Main Contributions}
\begin{description}\itemsep0.2em
\item[$C_1$] PSI-based quantification of non-IIDness in FL and comparison to alternative metrics.
\item[$C_2$] Clust-PSI-PFL: PSI-guided client clustering (K-means++) with silhouette-based cluster count selection and cluster-wise personalization.
\item[$C_3$] Empirical validation across 6 datasets and 2 partition protocols (Dirichlet($\alpha$), Similarity($S$)), reporting accuracy and fairness improvements vs.\ baselines.
\end{description}

\subsection{Computational Artifacts}
\begin{description}\itemsep0.2em
\item[$A_1$] Code repository (implementation + experiments):\\
\url{https://github.com/Sapienza-University-Rome/clust_psi_pfl}\\
\end{description}

\begin{table}[h]
\caption{Artifact-to-paper mapping (compact).}
\centering
\footnotesize
\setlength{\tabcolsep}{3pt}
\renewcommand{\arraystretch}{1.05}
\begin{tabularx}{\columnwidth}{p{0.11\columnwidth} p{0.20\columnwidth} X}
\toprule
\textbf{ID} & \textbf{Supports} & \textbf{Reproduces} \\
\midrule
$A_1$ & $C_1$--$C_3$ & Main experimental results (accuracy and fairness) and clustering behavior reported in the paper; see README.md for runnable configurations. \\
\bottomrule
\end{tabularx}
\end{table}

\section{Artifact Identification (A\texorpdfstring{$_1$}{1})}

\newartifact

\artrel
$A_1$ is the reference implementation of Clust-PSI-PFL and the baselines used in the evaluation (methods selectable via CLI). It enables (i) PSI-based non-IID quantification, (ii) PSI-driven clustering and $\tau$ selection, and (iii) federated training/evaluation on the supported datasets and partition protocols.

\noindent\textbf{License:} MIT. \quad

\artexp
Expected outcomes (trend-level):
(i) Stronger non-IID (smaller $\alpha$ / smaller $S$) yields worse baseline performance; (ii) Clust-PSI-PFL improves global accuracy and reduces client disparity (fairness); (iii) Clustering selects a small $\tau$ under the silhouette criterion. Exact numbers may vary due to nondeterminism and environmental differences.

\arttime
Setup: 10--30 min. Execution: 30--60 min (smoke test); paper-like sweeps: 26 hours, including all the baselines for the slower dataset, which is Sent140. Analysis: 10--20 min.

\artin
\artinpart{Hardware}
Validated on Linux with high-end CPU/RAM and an NVIDIA RTX A6000 (48GB). A CUDA GPU is recommended for image/text datasets; CPU-only reproduction is possible but slower. Disk: 1TB.

\artinpart{Software}
Linux; Python 3.10 via conda env \texttt{clust\_psi\_pfl\_py310}. Install all dependencies from \texttt{environment.yml}. Key versions: Follow the versions included in the \texttt{environment.yml} file.

\artinpart{Datasets / Inputs}
Supported datasets: \texttt{acs\_income, serengeti, fmnist, cifar10, sent140, amazon\_reviews}.
Sent140 must be downloaded manually (not shipped); follow the README.md instructions and expected folder layout under \texttt{data/sent140/}.
For the remaining datasets, no manual dataset folder preparation is required; in particular, \texttt{cifar10} and \texttt{fmnist} are obtained via standard dataset utilities (i.e., no repository data folder is needed in advance).

\artinpart{Installation and Deployment}
\begin{itemize}\itemsep0.2em
\item Clone repo, then create/activate conda env from \texttt{environment.yml}.
\item If running Sent140, download and place required files under \texttt{data/sent140/} (README.md).
\item It can be run on a single machine (GPU is optional but recommended, especially for image and text datasets).
\end{itemize}

\artcomp
Workflow (tasks/dependencies):
$T_1$ partition data (\texttt{--dataset}, \texttt{--partitioner}, \texttt{--non-iid-param}, \texttt{--num-clients})
$\rightarrow$
$T_2$ (Clust-PSI-PFL only) compute PSI features + cluster selection (silhouette) / fixed \texttt{--tau-clusters}
$\rightarrow$
$T_3$ federated training/evaluation (\texttt{--agg-method}) producing accuracy and fairness metrics.
For non-clustering baselines: $T_1 \rightarrow T_3$.
Paper-like settings: $T\!=\!40$ rounds, $E\!=\!5$ local epochs, 5 seeds/partitions; client fraction $q\!=\!0.5$.

\artout
Outputs include global accuracy, per-client local accuracy, and fairness metrics ($AD$, $SDAD$), printed.

\newpage
\appendixAE

\arteval{1}
\artin
\url{https://github.com/Sapienza-University-Rome/clust_psi_pfl/blob/main/README.md}

\noindent\textbf{Install/Run (minimal).}
\begin{itemize}\itemsep0.2em
\item Create env: \texttt{conda env create -f environment.yml}; activate: \texttt{conda activate clust\_psi\_pfl\_py310}.
\item Smoke test: \texttt{python main.py}.
\item If using Sent140: download into \texttt{data/sent140/} as per README.md.
\end{itemize}

\artcomp
\noindent\textbf{Reproduction procedure (compact).}
\begin{enumerate}\itemsep0.2em
\item Choose dataset and partition: Dirichlet (\texttt{--partitioner=dirichlet}, \texttt{--non-iid-param}=$\alpha$) or Similarity (\texttt{--partitioner=similarity}, \texttt{--non-iid-param}=$S$).
\item Run Clust-PSI-PFL: set \texttt{--agg-method=clust\_psi\_pfl}, and set rounds/epochs/seeds to paper-like values.
\item Run baselines by changing \texttt{--agg-method} (as listed in README.md), keeping the same partition and seeds. Baselines include FedAvg, FedProx, FedAvgM, FedOpt variants, and clustering/personalization baselines (see README.md).

\end{enumerate}

\noindent\textbf{Paper-level reproduction:}
for each dataset and non-IID setting, run 5 seeds/partitions, then compute mean$\pm$std
of (i) global accuracy and (ii) fairness metrics ($AD$, $SDAD$) over the 5 runs.

\begin{lstlisting}[basicstyle=\ttfamily\footnotesize,breaklines=true]
python main.py --dataset=acs_income --partitioner=dirichlet \
 --non-iid-param=0.3 --num-clients=100 --agg-method=clust_psi_pfl \
 --comm-rounds=40 --local-epochs=5 --seeds-list="42,0,1,2,3"
\end{lstlisting}

\noindent\textbf{Determinism:} results may vary due to GPU nondeterminism; fixing
\texttt{--seeds-list} minimizes variance (full determinism is not guaranteed).

\artout
\begin{itemize}\itemsep0.2em
\item \textbf{Expected results:} Under strong non-IID (small $\alpha$ / $S$), Clust-PSI-PFL should improve global accuracy and reduce client disparity vs.\ baselines (trend-level agreement).
\item \textbf{Evaluation:} Compare reproduced accuracy/fairness trends with the corresponding results in the paper (same dataset/partition). Acceptable tolerance: $\pm 1$ percentage point for global accuracy and
$\pm 5\%$ (relative) for fairness metrics ($AD$, $SDAD$); trend-level agreement
across non-IID settings ($\alpha$/$S$) is required.

\item \textbf{Contribution link:} PSI/non-IID trends support $C_1$; clustering behavior supports $C_2$; accuracy+fairness improvements support $C_3$.
\end{itemize}

\end{document}